\documentclass{article}
\usepackage{amssymb}
\usepackage{amsmath}
\usepackage{amsthm}
\usepackage{algorithm}
\usepackage{algpseudocode}
\usepackage{booktabs}
\usepackage{graphicx}
\usepackage{xcolor}
\usepackage{subcaption}
\usepackage{multirow} 
\usepackage{graphicx}
\usepackage{wrapfig}
\usepackage{enumitem}
\usepackage{placeins}
\usepackage{float}

\usepackage[preprint]{corl_2025} 

\title{TReF-6: Inferring Task-Relevant Frames from a Single Demonstration for One-Shot Skill Generalization}

\author{
  Yuxuan Ding
  \aspace Shuangge Wang
  \aspace Tesca Fitzgerald\\
  Yale University\\
\texttt{eason.ding@aya.yale.edu} \\
\texttt{\{shuangge.wang, tesca.fitzgerald\}@yale.edu}
}

\newcommand{\aspace}{\hspace{1em}}

\definecolor{darkgreen}{RGB}{0,128,0}
\definecolor{darkred}{RGB}{200,0,0}

\begin{document}

\maketitle

\begin{abstract}
    Robots often struggle to generalize from a single demonstration due to the lack of a transferable and interpretable spatial representation. In this work, we introduce \textbf{TReF-6}, a method that infers a simplified, abstracted \textbf{6}DoF \textbf{T}ask-\textbf{Re}levant \textbf{F}rame from a single trajectory. Our approach identifies an influence point purely from the trajectory geometry to define the origin for a local frame, which serves as a reference for parameterizing a Dynamic Movement Primitive (DMP). This influence point captures the task's spatial structure, extending the standard DMP formulation beyond start-goal imitation. The inferred frame is semantically grounded via a vision-language model and localized in novel scenes by Grounded-SAM, enabling functionally consistent skill generalization. We validate TReF-6 in simulation and demonstrate robustness to trajectory noise. We further deploy an end-to-end pipeline on real-world manipulation tasks, showing that TReF-6 supports one-shot imitation learning that preserves task intent across diverse object configurations.

\end{abstract}

\section{Introduction}

Robots are increasingly expected to operate in dynamic, human-centered environments, whether assisting in homes~\citep{intelligence2025pi05visionlanguageactionmodelopenworld}, collaborating in warehouses~\citep{warehouseKleerHRI}, or supporting kitchen workflows~\citep{shi2023robocook}. These settings demand generalizable behavior: adapting motion to unseen objects, adjusting to new object placements and orientations, and aligning with different surface orientations. While humans perform such adaptations effortlessly, robots struggle to generalize skills from limited demonstrations.

Since training data can never fully prevent out-of-distribution (OOD) scenarios, researchers increasingly turn to extract structural representations, such as rigid-body poses~\citep{wen2024foundationposeunified6dpose} or keypoints~\citep{manuelli2019kpamkeypointaffordancescategorylevel, liu2024mokaopenworldroboticmanipulation, huang2024rekepspatiotemporalreasoningrelational}, which provide a more stable foundation for generalization in manipulation tasks. Recent work has developed generalizable techniques for generalizing a task in terms of goal poses~\citep{li2025elasticmotionpolicyadaptive, pan2022taxpose, se3representation2022}, sub-goals~\citep{subgoalOchoa2024}, or contact interactions~\citep{oneShotVisualContactRich}. There has been less work, however, on generalizing the spatial constraints encoded in the trajectory itself, such as the curvature of opening the door indicating a hinge constraint besides just the handle position. In practice, the shape of a human demonstration reflects more than just start and end points. It encodes implicit constraints, such as obstacle avoidance~\citep{obstacleWinn2012}, mechanical limitations (e.g. hinge constraints)~\citep{constraintGuru2018}, or ergonomic preferences~\citep{humanPreferenceBestick2018}. For any two points in space, infinitely many paths exist, yet demonstrators tend to follow specific, repeatable curves. Prior works in~\citep{geometricJun2020} have shown that humans opt for these repeatable trajectories due to their similarities in more condensed, latent geometric structure. We believe that inferring these latent structures could be informative in generalizing trajectories to unseen objective configurations.
%
Furthermore, we expect that they may correspond to semantic features that are detectable by vision-language models (VLMs)~\citep{zhang2025iaaointeractiveaffordancelearning}.
This motivates our central research question: \textit{Can we use a single demonstration to infer a task-relevant spatial reference frame that is} \textbf{(1)} \textit{semantically identifiable (anchors to scene features)} and \textbf{(2)} \textit{functionally meaningful (preserves constraints), enabling generalization across object poses and configurations?}

\begin{figure}[t]
    \centering
    \includegraphics[width=\linewidth]{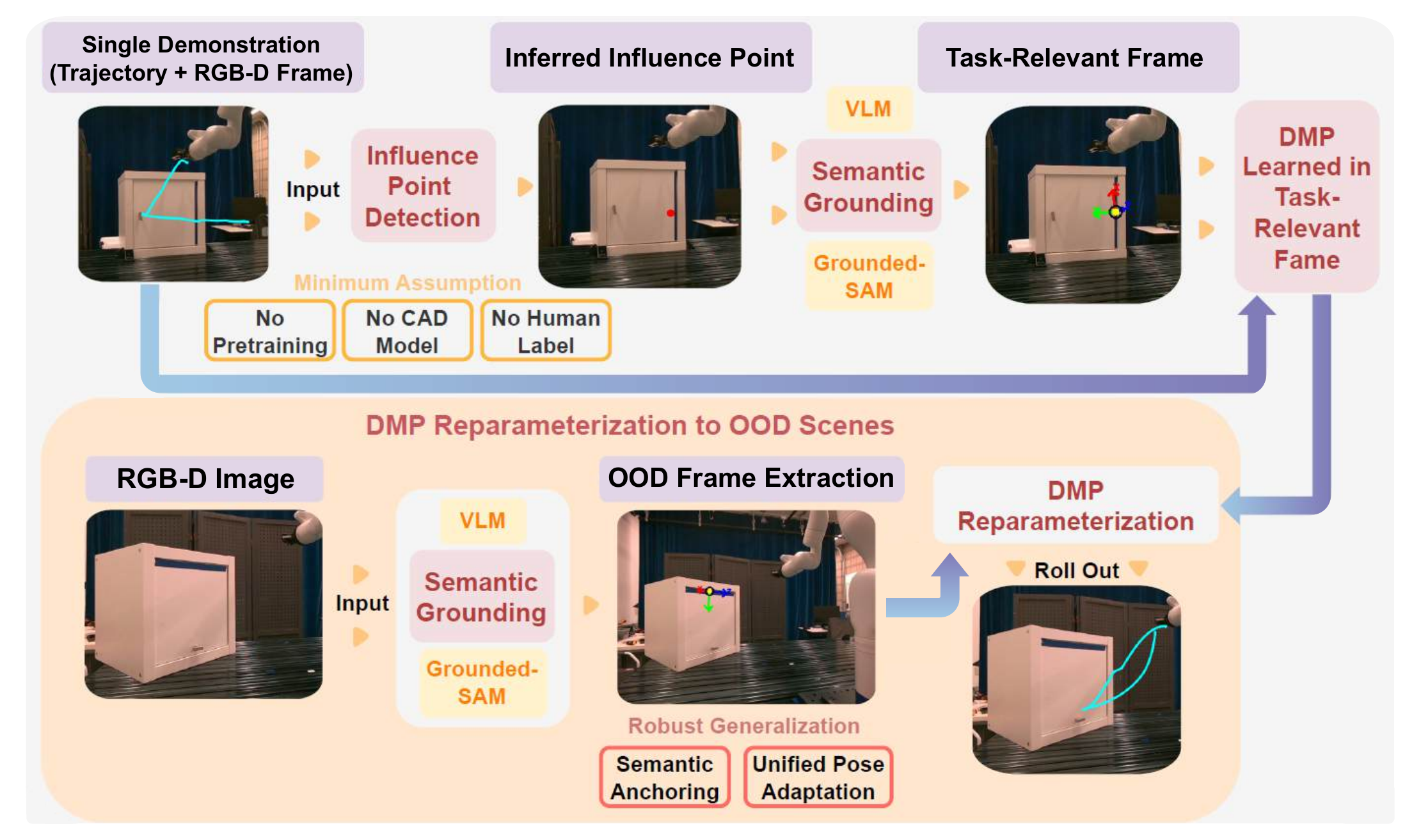}
    \caption{\textbf{Overview of TReF-6.} 
    Given a single demonstration, TReF-6 infers an implicit influence point, semantically grounded by a vision-language model (VLM), and extracts a 6-DoF reference frame from the segmentation provided by Grounded-SAM. With minimum assumptions, the inferred frame enables robust OOD generalization.}
    \label{fig:pipeline}
\end{figure}

In this work, we propose \textbf{TReF-6} (Figure~\ref{fig:pipeline}), a framework that infers a task-relevant, 6DoF reference frame from a single demonstration that enables simple dynamical controllers such as DMPs to generalize their motion robustly. 
Our contributions include:

\begin{enumerate}[leftmargin=*]
    \item Formalizing the problem of inferring a task-relevant frame as an optimization over its geometric consistency to the trajectory dynamics;
    \item An efficient optimization algorithm that is robust to trajectory noise and does not rely on object priors, human labels, or dense annotations; and
    \item Simulated and physical robot evaluations that demonstrate our method's ability to generalize to new spatial variations of demonstrated tasks. 
\end{enumerate}

\section{Related Work}

\textbf{Dynamic Movement Primitives} (DMPs) encode a demonstrated trajectory as a stable dynamical system~\citep{ijspeert2002movement, ijspeert2013dynamical} that converges toward a goal. 
DMPs can generalize the shape of a demonstrated motion to target a new goal pose, yet are functionally agnostic to the surrounding environment unless manually augmented with additional force-like terms that account for external influences~\citep{pastor2009learning} or goal and sub-goal related task frames~\citep{frameDMP}. Without access to an appropriate task-specific reference, DMPs are trained over just a single trajectory, so by definition they ``overfit" to the environmental configuration and constraints that were present during the initial demonstration. 

\textbf{Data-driven approaches} such as TP-GMMs~\citep{Calinon16JIST} and their extensions~\citep{huang2018generalizedtaskparameterizedskilllearning} involve conditioning a demonstrated trajectory on external task parameters. These approaches improve adaptation across different object poses, placements, and orientations, but require extensive training data (including synthetic data) to generalize effectively~\citep{Calinon16JIST, zhu2022learningtaskparameterizedskillsdemonstrations}, or assume that relevant task-relevant frames, such as those aligned with object or interaction axes, are provided or can be easily extracted~\citep{hu2019hierarchical}. 


\textbf{Affordance learning} focuses on the actionable possibilities in an environment, such as regions on an object that can be grasped, pushed, or opened~\citep{gibson2014ecological}, capturing the physical interactions that enable task completion. Recent efforts in this field have focused on visually identifying object surfaces that support affordances based on robot-collected data~\citep{yenchen2022miramentalimageryrobotic} or human demonstration videos~\citep{bahl2023affordanceshumanvideosversatile, chen2023affordancegroundingdemonstrationvideo}. Others use depth data to address occlusion and collision constraints in articulated objects~\citep{cheng2023learningenvironmentawareaffordance3d}. While these approaches have shown promising results, they involve training visual backbones with target object types or multi-view RGB inputs. In real-world scenarios where a robot must interact with unseen novel objects, especially under varying camera viewpoints with crowded objects and random occlusions, such reliance becomes a bottleneck in terms of data requirements and system robustness.

\textbf{One-shot imitation learning} targets this challenge of adapting to novel task constraints or objects. This may involve grounding the demonstration using external context—language descriptions or videos of related tasks to make sense of the action in a broader setting~\citep{BC-Z}. Other methods focus on low-level structure, aligning object parts or motion trajectories across different scenes to find transferable patterns~\citep{Hadjivelichkov2022affcorrs, li2025elasticmotionpolicyadaptive, vitiello2023oneshotimitationlearningpose} or with rich contact information~\citep{oneShotVisualContactRich}. Recent work also looks for structure that is consistent across different environments, for example, regions in 3D space that consistently guide behavior across tasks~\citep{InvariantRegionIn3D}, or prior demonstrations retrieved based on graph-based similarity metrics~\citep{yin2024offline}. These approaches enable a robot to quickly generalize its task knowledge to novel constraints and objects, and reinforce the importance of a reasoning module that correctly and efficiently identifies task-relevant features for one-shot learning~\citep{ajay2023compositionalfoundationmodelshierarchical}.

Although efficient at generalization, one-shot methods still require a large amount of data at training to acquire the ability to learn-to-learn and generalize from a single new example. Compared to end-to-end one-shot policy learning approaches, DMPs offer advantages in stability, data efficiency, and ease of adaptation once a transferable frame is available. This motivates our focus on building a minimal-assumption method to automatically extract a task-relevant spatial frame, enabling classical DMPs to generalize motion trajectories without the need for large-scale task distributions, external labels, or object CAD models.

\section{Methodology}

%


We introduce \textbf{TReF-6}\footnote{\href{https://github.com/iqr-lab/tref-6}{GitHub repository: https://github.com/iqr-lab/tref-6}}, a trajectory-based framework for inferring a task-relevant local frame from a single demonstration for skill generalization. Unlike prior work that relies on large-scale training data or assumes external knowledge about object geometry, our method leverages a novel optimization-based formulation to infer latent influence point from motion dynamics, which is general-purpose and operates on a single trajectory. The end-to-end pipeline consists of three stages:
\begin{enumerate}[leftmargin=*]
    \item \textbf{Influence Point Inference:} Optimize a directional consistency score to identify a 3D spatial point that best captures the trajectory's dynamics.
    \item \textbf{Semantic Grounding:} Refine the point by aligning it with a semantically relevant and spatially aligned visual feature identified by a VLM, and extract a local frame based on surface normals and interaction direction.
    \item \textbf{DMP Reparameterization:} Transform the trajectory into the inferred local frame, fit DMPs, and reuse them in novel scenes by extracting a new frame.
\end{enumerate}

\subsection{Influence Point Inference: Defining a Directional Consistency Score}

We formulate the frame inference problem as identifying a coordinate transformation informed by the trajectory. We hypothesize that the demonstrated motion is shaped by a dominant spatial constraint like rotation around a hinge (axis constraint), movement along a shelf (plane constraint), or alignment toward a socket (point constraint). Rather than designing separate models for each type, we infer a single 6-DoF frame centered at a latent point $p\in \mathbb{R}^3$ that reflects the underlying structure influencing the motion, serving as a general-purpose reference for motion generalization.

Given a single trajectory $\{x_t\}^T_{t=1} \in \mathbb{R}^3$, we hypothesize that the motion can be captured by a position-only, latent influence pointing from $x_t$ to a fixed point $p \in \mathbb{R}^3$. We optimize $p$ for both \textit{temporal consistency} and \textit{directional agreement}, whether the observed acceleration aligns consistently toward a candidate point, inspired by prior work in shared-control robotics to disambiguate user intent~\citep{9066939}. Specifically, we define the \textbf{directional consistency score}, $\mathcal{S}: \mathbb{R}^3 \rightarrow \mathbb{R}$, which compares the predicted direction from $x_t$ to $p$ with the acceleration, $\Ddot{x}_t$. We define $\mathcal{S}(p)$ as:
\label{eq:force_residual_score}
\begin{equation}
    \mathcal{S}(p) = -\frac{1}{T}\sum^T_{t=1}
    \left|\left|\frac{p -x_t}{||p - x_t|| + \epsilon} - \Ddot{x}_t\right|\right|
\end{equation}
where $\epsilon > 0$ is a small positive constant. A higher $\mathcal{S}(p)$ value indicates stronger consistency between the trajectory's underlying dynamics and the candidate point $p$. A 2D example of the score landscape is shown in Appendix Figure \ref{fig:high_score}, where regions around the ground-truth candidate point of influence have higher $\mathcal{S}(p)$ values. By relying on the second derivative, the score inherently reflects the temporal dynamics of the trajectory. Since direction difference is normalized, $\mathcal{S}$ provides an objective that is robust to variations in force magnitude.

\subsection{Influence Point Inference: Optimizing for Directional Consistency}
\label{sec:landscape}

We can then estimate the latent intent point by solving for $p^* = \arg\max_{p \in \mathbb{R}^3}\mathcal{S}(p)$.
Due to vector normalization and temporal error aggregation, we find that the score landscape exhibits many local optima and extensive flat regions, especially under noise in acceleration direction, as illustrated in Appendix Figure \ref{fig:non-convex}, making the result highly sensitive to initialization. Empirically, we find that choosing an initialization point with $(1)$ \textit{sharp gradients} and $(2)$ \textit{proximity to the ground truth}, is essential for successful optimization - an insight aligned with prior works in optimization~\citep{Cand_s_2015, Chi_2019}. Formally, if we denote the small angular deviation between $\frac{p-x_t}{||p-x_t||}$ as $\theta$, we can apply the law of cosines to approximate the squared residual and its partial derivative as follows:
\begin{align}
    \left|\left|\frac{p-x_t}{||p-x_t|| + \epsilon} - \Ddot{x}_t\right|\right|^2 &\approx ||\Ddot{x}_t||^2-2||\Ddot{x}_t||\cos{\theta} + 1 \\
    \implies \frac{\partial}{\partial \theta} \left|\left|\frac{p-x_t}{||p-x_t|| + \epsilon} - \Ddot{x}_t\right|\right|^2 &\approx \frac{\partial}{\partial \theta} \left(||\Ddot{x}_t||^2-2||\Ddot{x}_t||\cos{\theta} + 1\right) = 2||\Ddot{x}_t||\sin{\theta}
\end{align}
%
%
where the change in residual magnitude due to the deviations in $\theta$ is proportional to $||\Ddot{x}_t||$. 
We initialize the optimization near these regions by computing $p_0 \sim  \frac{1}{k} \sum_{\tau \in \mathcal{I}}x_j + \mathcal{N}(0, \sigma^2\mathbf{I})$,
where $\mathcal{I}$ is the top$-k$ timesteps with the largest $||\Ddot{x}_t||$ magnitudes.
%
This initialization places the optimization near points of sharper gradients. We empirically show in Section \ref{sec:sim_experiment} that this approach significantly improves convergence and solution quality. Due to the non-smoothness of normalization near $p\approx x_t$, we approximate gradients using central finite differences and use Adam~\citep{kingma2014adam} to smooth fluctuations in the gradient and accelerate convergence. Algorithm~\ref{alg:fitting} details the entire optimization algorithm.

\begin{algorithm}[t]
\caption{Trajectory-based Influence Point Identification}
\label{alg:fitting}
\begin{algorithmic}[1]
\Require Trajectory positions $\{x_t, \Ddot{x}_t\}_{t=1}^T$, learning rate $\eta > 0$, total steps $N$, initialization count $k$
\State $\mathcal{I} = \arg\max_{\substack{\mathbb{J} \subseteq \{1, \dots, n\} \\ |J| = k}} \sum_{j\in \mathbb{J}} ||\Ddot{x}_t||$ \Comment{Select top-$k$ time steps with highest norms}
\State Sample $ p \sim  \frac{1}{k} \sum_{\tau \in \mathcal{I}}x_j + \mathcal{N}(0, \sigma^2\mathbf{I})$ \Comment{Sample Initial State}
\For{$i = 1$ to $N$}
    \State 
        $\nabla_p \mathcal{S}(p) \approx \left[ \frac{\mathcal{S}(p + \epsilon e_i) - \mathcal{S}(p - \epsilon e_i)}{2\epsilon} \right]_{i=1}^{3}$ \Comment{Estimate Gradient}
        
    \State $p \leftarrow p - \eta \nabla_p \mathcal{S}(p)$  
\EndFor
\State \Return $p$
\end{algorithmic}
\end{algorithm}

\subsection{Semantic Grounding}

\begin{figure}[t]
    \centering
    \includegraphics[width=1.0\linewidth]{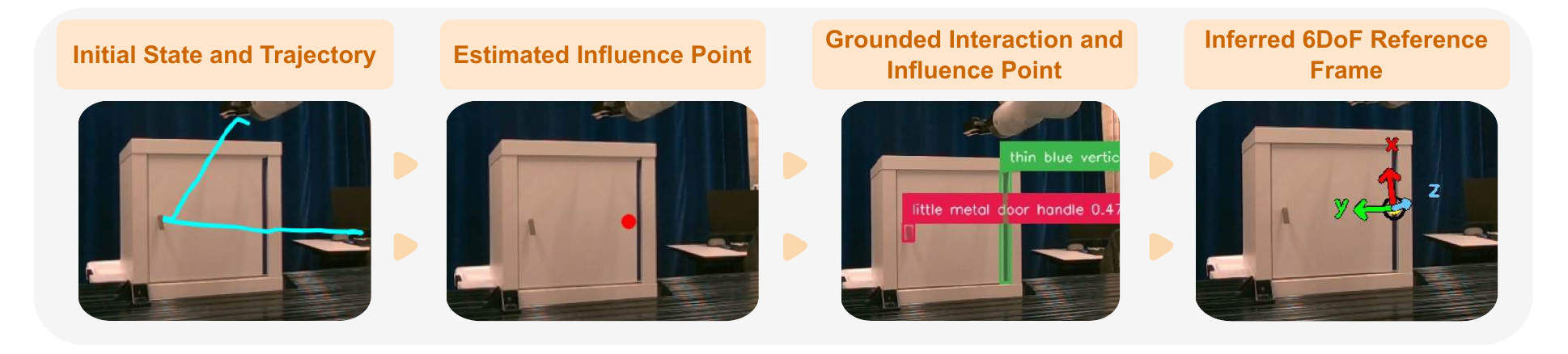}
    \caption{6DoF Frame Extraction for Door Opening Demonstration.}
    \label{fig:frame_extraction}
\end{figure}

We then use the optimized influence point $p^*$ as the origin of the task-relevant 6DoF Frame. To ensure the frame is semantically meaningful and transferable across scenes, we refine this point by aligning it with visual features identified by GPT-4o~\citep{openai2024gpt4o}. Specifically, we implement a two-phase querying process. In the first phase, we prompt the model to generate a high-level task label based on the initial state RGB image overlaid with the demonstration trajectory. In the second phase, using the predicted task label, we query GPT-4o again with the same RGB image now overlaid with the inferred influence point to identify the visual features associated with both the influence point and the interaction point (where the robot first begins its interact with the environment). We provide the full prompts in Appendix \ref{sec:prompt}. The model returns a natural language description of the features, which we use to guide segmentation via Grounded-SAM and refine the point.
Once the refined point is established, we compute the surface normal at that location to define the $z$-axis of the frame. We then define the $yz$-plane using the $z$-axis and the vector from the refined point to the interaction point. The $x$-axis is then computed as the unit vector orthogonal to this plane. Finally, the $y$-axis is obtained as the cross product of the $z$- and $x$-axes. This captures both local geometry (via the surface normal) and task-relevant directionality (via the interaction point), as shown in Figure~\ref{fig:frame_extraction}.

\subsection{DMP Reparameterization}

To enable environment-adaptive generalization, we apply the standard DMP formulation in a task-relevant frame centered at the refined influence point $p^*$. Rather than modifying the DMP dynamics, we reparameterize the demonstrated trajectory by transforming it from the robot base frame, referred to as the world frame, into the task-relevant frame.
Given a demonstration of length $T$, the Cartesian position $x_t$ and quaternion orientation $q_t$ at time $t$ of the trajectory are transformed from the world frame into the inferred task-relevant frame.
We then compute the relative motion from the starting pose in this local frame:
\begin{equation}
\begin{aligned}
    \Delta x_t &= x^{\text{local}}_t - x^{\text{local}}_0 \qquad\qquad\qquad\qquad
    \Delta q_t &= q^{\text{local}}_t \otimes q_0^{\text{local}^{-1}}
\end{aligned}
\end{equation}
where $x_0, q_0$ denote the initial end-effector pose relative to the influence point, and $\otimes$ is quaternion multiplication. This initial pose corresponds to the location of the tool at the beginning of task execution (e.g., where a brush first makes contact with the robot). We then fit Cartesian and quaternion DMPs over $\Delta x_t$ and $\Delta q_t$, respectively.
At deployment, a new influence point $p^*$ and associated task-relevant frame are inferred, and the new starting pose $(x^{\text{new}}_0, q^{\text{new}}_0)$ is computed accordingly. The DMPs then roll out relative motions with respect to $(x^{\text{new}}_0, q^{\text{new}}_0)$. The definition of the roll out functions $\text{DMP}_{\text{pos}}(x^{\text{new}}_0)$ and $\text{DMP}_{\text{quat}}(q^{\text{new}}_0)$ is deferred to Appendix \ref{sec:dmp}.
%
%
The generated trajectory is finally mapped back into the world frame using the inverse transformation.
\section{Experiment}

To evaluate whether our proposed method could infer a task-relevant spatial reference frame that is (1) semantically identifiable and (2) functionally meaningful from a single demonstration, we designed controlled 3D simulations with known influence point to assess inference precision under varying levels of directional and magnitude noise in acceleration, using Mean Euclidean Distance Error (MEDE) between the inferred and ground-truth influence point as the metric. In real-world tasks, we assess whether the inferred frame can be semantically grounded and enable one-shot skill generation. While most sophisticated baselines such as affordance-based imitation or goal-conditioned policies exist, they typically rely on object CAD models~\citep{li2025elasticmotionpolicyadaptive}, extensive training~\citep{InvariantRegionIn3D}, or rich contact information~\citep{oneShotVisualContactRich}, which are unavailable in our setting. We thus benchmark against privileged DMPs, which have access to additional information such as object positions or operate in environments that mimic the demonstration setup, defined in Section \ref{sec:privileged}, as the strongest feasible baseline. Improvement over the baseline, measured by task success under an OOD environment, indicates that the inferred frame captures structure critical for generalizing the demonstrated behavior.

\subsection{Simulated Environment}
\label{sec:sim_experiment}

In real-world settings, demonstrations are inherently noisy, and our method relies on the second-order trajectory, which amplifies noise. Therefore, to test the robustness of our method with noisy demonstrations, we design a 3D simulation where the ground-truth influence point and motion dynamics are fully controllable. A point mass, simulating a robot end-effector, starts at the origin with a random velocity sampled from $[-0.5, 0.5]^3$, and is influenced by a randomly placed, ground-truth influence point $\hat{p}$ within a bounded region $[-5, 5]^3$. A directionally noisy force points from the particle to the influence, scaled by a coefficient $\alpha_t \sim \mathcal{U}(0, ||\hat{p} - x_t||)$, mimicking diminishing attraction. We add a constant Gaussian noise to both direction and magnitude of control. The state evolves throughout a locally linearized dynamics for $100$ steps.


We evaluated our method with and without random initialization along with three other scoring objectives and one inverse dynamics baseline for inferring the spatial influence point $p$. All methods were tested under the same simulated conditions. We reported the MEDE between the predicted and ground-truth influence points over $50$ randomized seeds. These baselines were selected to represent diverse but plausible strategies for trajectory-based influence point inference without requiring access to ground-truth labels, including a physically grounded method (inverse dynamics triangulation), optimization over residual scores (quadratic residual score), and directional alignment (cosine similarity score).

\begin{figure}[t]
    \centering
    \includegraphics[width=\linewidth]{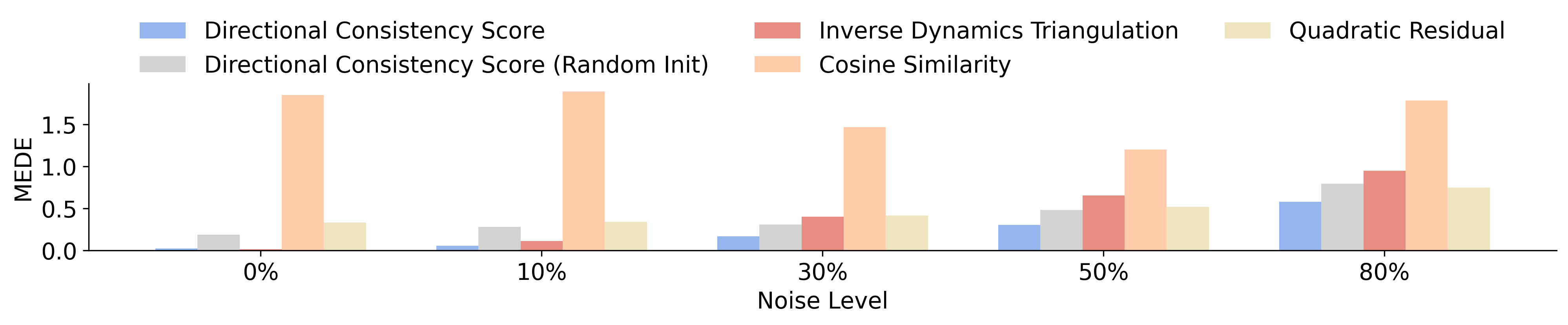}
    \caption{Mean Euclidean Distance Error (MEDE) comparison of spatial influence inference methods under varying levels of noise (0\% to 80\%). Complete results are summarized in Appendix \ref{sec:simulation_more}}
    \label{fig:influence_comparison}
\end{figure}

\begin{figure}[t]
    \centering
    \begin{minipage}{\textwidth}
        \centering
        \includegraphics[width=\linewidth]{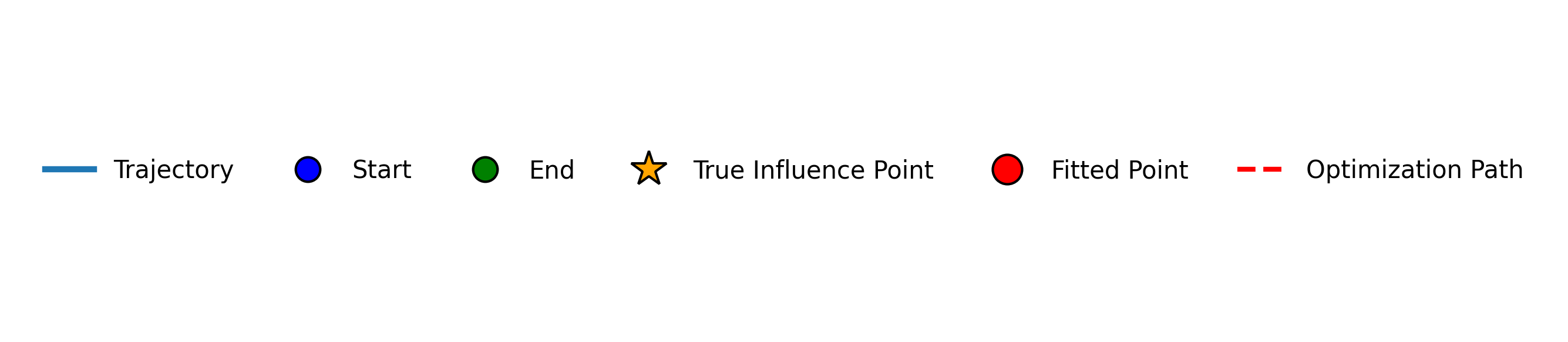}
    \end{minipage}
    \vspace{0.5em}
    \begin{subfigure}[t]{0.48\textwidth}
        \centering
        \includegraphics[width=\linewidth]{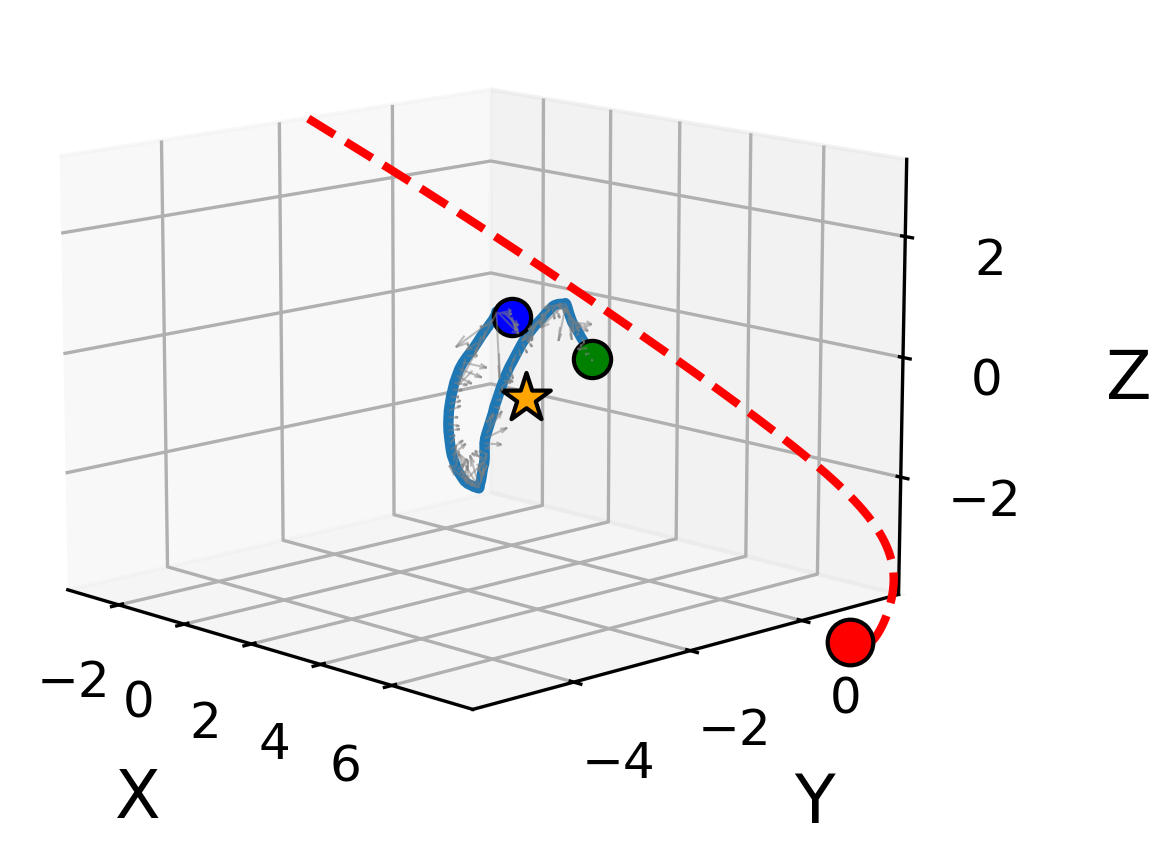}
        \label{fig:failure_random_init}
    \end{subfigure}
    \hfill
    \begin{subfigure}[t]{0.48\textwidth}
        \centering
        \includegraphics[width=\linewidth]{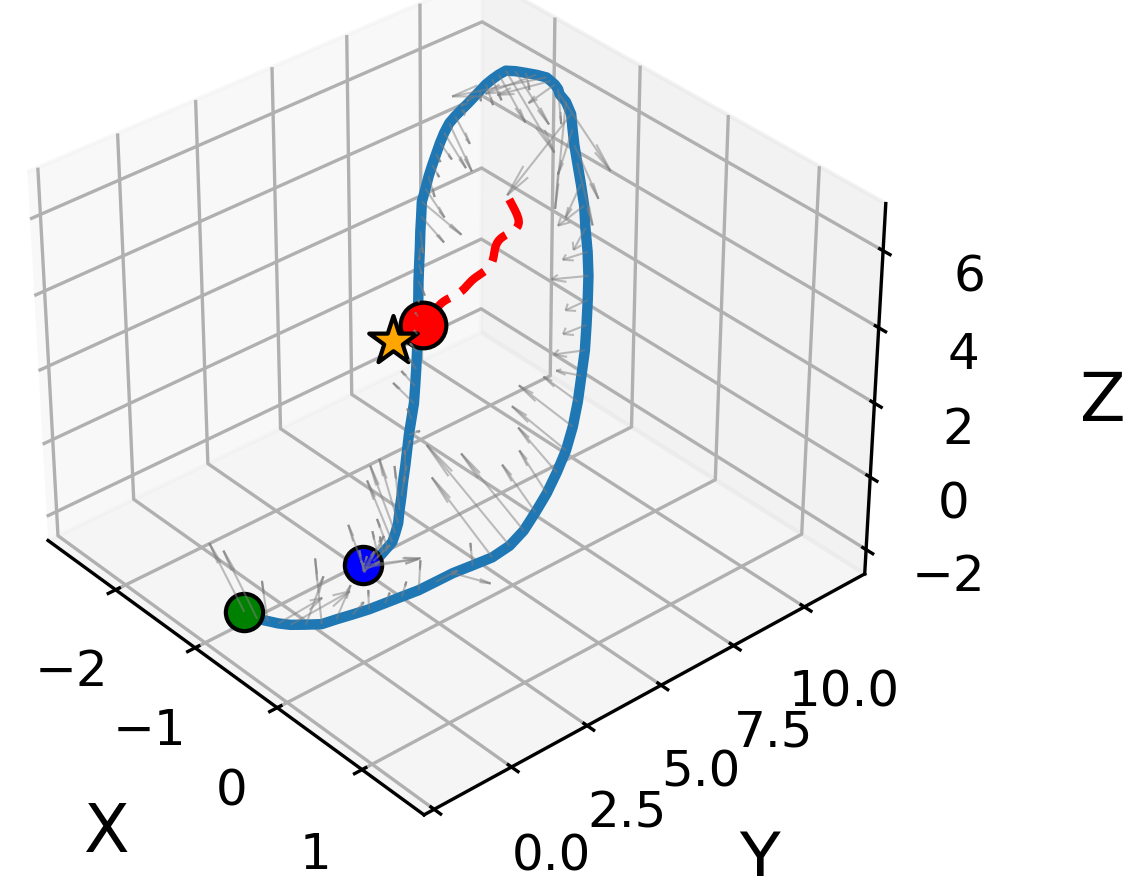}
        \label{fig:failure_structured_init}
    \end{subfigure}
    \caption{\textbf{Qualitative comparison between random and structured initialization under 50\% noise.}  Gray arrows represent $\ddot{x}$, which appear noisy due to $50\%$ noise level, and the dashed red line illustrates the optimization trajectory. (Left) Random initialization is trapped in a flat-gradient region. (Right) Structured initialization converges close to the true influence despite noisy acceleration.}
    \label{fig:initialization}
\end{figure}

\textbf{Directional Consistency Score (ours)} achieved the best overall performance across all noise levels, maintaining low error and variance even as the force signal became increasingly corrupted, consistently outperforming random initialization baseline across all noise settings, with an average of $55.0\%$ reduction in error and an average of $86.8\%$ reduction in variance. Appendix Figure \ref{fig:initialization} illustrates a representative worst-case outcome under $50\%$ noise using random initialization. In contrast, across all baselines, we observed distinct failure modes under extreme noisy conditions. \textbf{Inverse Dynamics Triangulation} was highly sensitive to perturbations in the direction of the force vector, as it relied on accurate ray intersections from $\Ddot{x}_t$ orientations. \textbf{Cosine Similarity Score} suffered in 3D settings due to vanishing gradients, leading to poor convergence and high variance. \textbf{Quadratic Residual Score} broke down when the force magnitudes were irregular or noisy—as was the case in our simulations where magnitudes were modulated stochastically. In contrast, our proposed \textbf{Directional Consistency Score} remained robust by comparing normalized force predictions directly with observed accelerations, achieving the best performance across varying noise levels.

Figure \ref{fig:influence_comparison} summarizes these results. Mathematical formulations of Cosine Similarity Score, Quadratic Residual Score, and Inverse Dynamics Triangulation are summarized in Appendix \ref{app:surrogate}. More detailed analysis and discussion of each approach as well as a comprehensive simulation experiment results are summarized in Appendix \ref{sec:simulation_more}.

\subsection{Real-World Experiment}

Our real-world experiments aim to validate that \textbf{TReF-6} enables functionally meaningful one-shot skill generalization, even when paired with simple downstream controllers such as DMPs. We evaluate its effectiveness across three real-world manipulation tasks on a 7-DoF Kinova Gen3 robot where trajectory shape and alignment are critical: (1) \textit{peg-in-hole dropping}, (2) \textit{cabinet door opening}, and (3) \textit{surface wiping}. Each task includes one demonstration and multiple OOD variants for evaluation. The peg-in-hole task tests for semantic transferability and precision, with variations in object shape and color, as well as rod color and height. The door-opening task examines generalization across spatial positioning and rotations, varying cabinet position, hinge placement, and cabinet orientation. The wiping task evaluates adaptability to surface tilt and the ability to maintain continuous surface contact without excessive force, with changes in stain appearance and the flatness of the board. 

\label{sec:privileged}
To isolate the contribution of our method, we adapt DMPs as the shared motion controller and compare executions with and without our inferred local reference frame. Since vanilla DMPs lack semantic grounding, we provide a privileged setup for baseline DMPs: objects and rods in the peg-in-hole task are placed in the original demonstration locations; for door opening, the handle position is explicitly specified; and for wiping, the whiteboard brush and tilted surface are arranged to match the demonstration.

\begin{wrapfigure}{r}{0.5\textwidth}
    \vspace*{-5mm}
    \centering    \includegraphics[width=\linewidth]{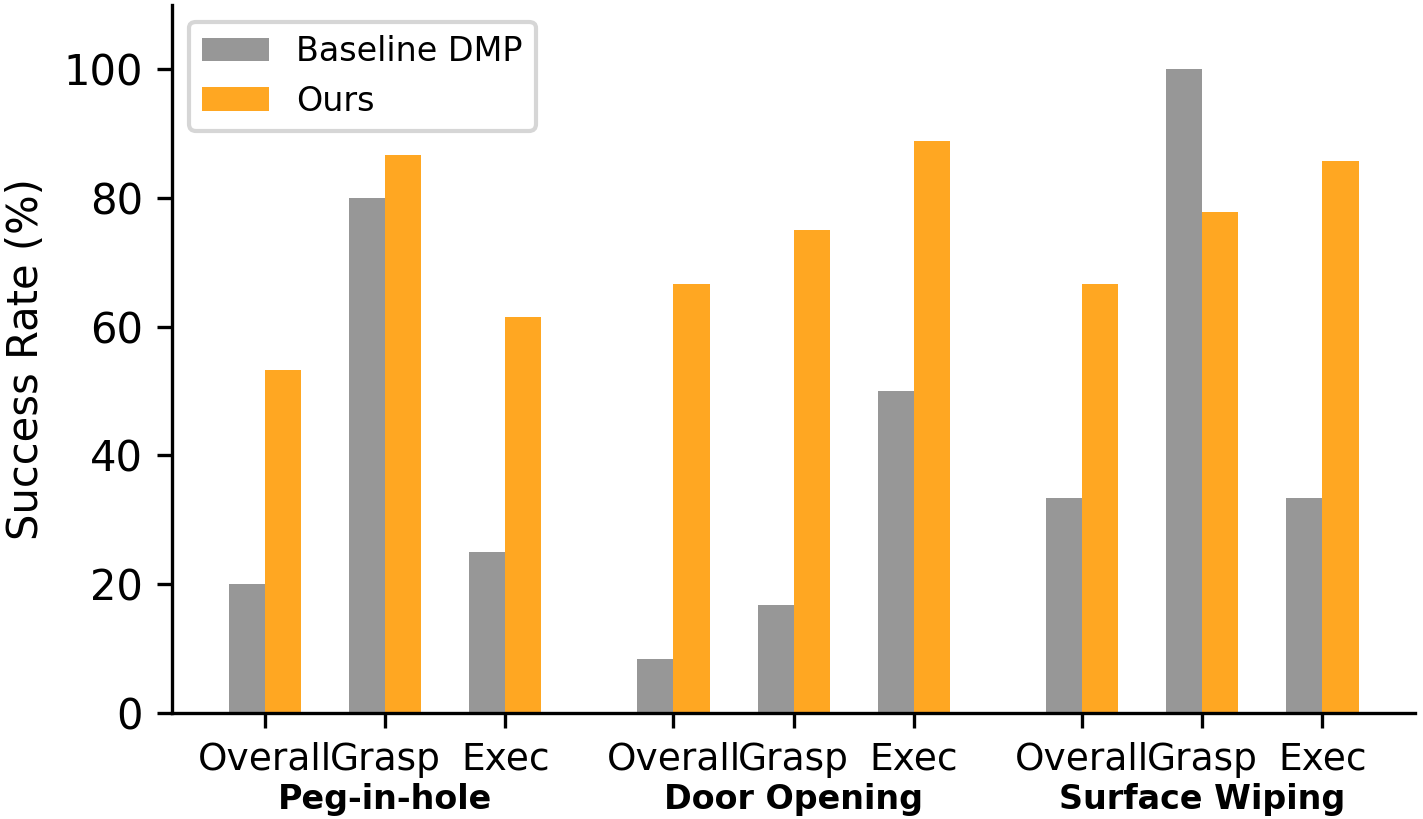}
    \caption{Real world experiment results. Each task includes: Overall success, Grasp success, and Execution success given successful grasp. Complete results were summarized in Appendix \ref{sec:real_world}}\label{fig:task_success_rates}
    \vspace*{-5mm}
\end{wrapfigure}

    \textit{(1) Peg-in-hole Dropping:} In this task, the robot is provided with a single demonstration, illustrated in Appendix Figure \ref{fig:demonstration}, and both methods successfully reproduce the demonstrated behavior in the original setting. We intentionally designed the task to be challenging and tightly constrained: as illustrated in Appendix Figure \ref{fig:rod_height_comparison}, slight misalignments between the object and the rod lead to failure. Our method maintains a high success rate under variations, as shown in Figure \ref{fig:task_success_rates}. While the baseline achieves a similar success rate in the pickup phase due to its privileged setup, its performance in completing the full task, assuming the object has been picked up, is significantly lower than ours.

    \textit{(2) Cabinet Door Opening:} The robot is provided with a single demonstration, illustrated in Appendix Figure \ref{fig:demonstration}, where it opens a cabinet door directly in front of it, left to right. Successful task execution requires rotating the handle around a hinge axis - a simple linear pulling would result in failure, as it would drag the entire cabinet without opening the door. Both the baseline and our method reproduce the demonstrated behavior in the original setting. However, DMP fails to adapt its motion to the new hinge geometry or cabinet orientation. In contrast, our method generalizes the motion successfully across multiple variants with an overall success rate of $66.7\%$ as shown in Figure \ref{fig:task_success_rates}.

    \textit{(3) Surface Wiping:} In this task, the robot is provided with a single demonstration, illustrated in Appendix Figure \ref{fig:demonstration}. A successful execution requires wiping out the ink on the whiteboard while maintaining contact with the surface, without deforming it to the point of touching the threshold placed behind. Both the baseline and our method reproduce the demonstrated behavior under the original setup. As shown in Appendix Table \ref{tab:success_rates_wiping}, the baseline performs reliably in this setting and under stain color changes. However, its motion does not adapt to different surface orientations. In contrast, our method achieves a much higher overall success rate of $66.7\%$ across tilted surface variants. When evaluating only trails where the brush was successfully grasped, our method achieves $85.7\%$ success, outperforming the baseline by $52.4\%$.

\section{Discussion and Analysis}


Overall, our real-world evaluations show that \textbf{TReF-6} enables functionally meaningful generalization across spatial variations with only a single demonstration by providing a semantically transferable task-relevant frame. We now highlight the key takeaways of our results.

\paragraph{Task-Specific Strength.}

Across all three tasks, \textbf{TReF-6} exhibits consistent improvements over baseline DMPs by leveraging task-specific spatial cues. In the \emph{peg-in-hole dropping} task, it adapts to changes in rod height and color, successfully adjusting the hook trajectory, while baseline DMP fail to hook the rod securely before release. In the \emph{door opening} task, \textbf{TReF-6} correctly infers the arc direction and hinge orientation, unlike baseline which follows a fixed left-to-right arc. In the \emph{surface wiping} task, \textbf{TReF-6} realigns motion to maintain contact to different tilted surfaces, which the baseline cannot handle.

\paragraph{Partial Success Despite Failures.}

Even when task execution is not fully successful, \textbf{TReF-6} often produces structurally meaningful motions. Failures in peg-in-hole dropping tasks arise primarily from color misclassifications (e.g., identifying red or blue rods as a purple rod), yet the behavior remains robust enough to complete partial goals (e.g. reaching the wrong rod). Most failures in door opening task arise from unreachable grasp pose, particularly when the handle is located near the table surface, leading the robot to hit kinematic limits. When grasping fails, the resulting motion still preserves arc structure around the hinge, reflecting meaningful generalization without success in the full task.

\paragraph{Performance Correlates to Perception Accuracy.}

The quality of generalization is tightly coupled with the reliability of depth perception. One example frame extraction failure case of dropping on the green rod that caused by depth noise is visualized in Appendix Figure \ref{fig:drop_failure}. Failure in the mirrored variant of door opening task is also attributed to poor depth perception when the door surface is nearly orthogonal to the camera, as shown in Appendix Figure \ref{fig:door_frame_mirror}. The lowest performance on surface wiping occurs at the $30^\circ$ tilt, as shown in Appendix Figure \ref{fig:wiping_frame_comparison}. We hypothesize that shallow tilts,  oblique angles, and small cross-section area of object lead to noisier or incomplete depth maps, leading to frame extraction failure. However, these perception challenges, such as segmentation errors or depth noise, may be mitigated in the future through advances in 3D perception and improvements to grounding models like Grounded-SAM, especially considering that we did not fine-tune the model for our tasks.
\section{Conclusion}

We presented a novel framework to augment DMPs' generalization capabilities: inferring a task-relevant spatial reference frame from a single demonstration that is both semantically identifiable and functionally meaningful. Unlike prior approaches that rely on object priors, multiple demonstrations, or predefined frames, our method extracts a full 6-DoF reference frame directly from the geometry of the demonstrated trajectory. By anchoring this inferred frame to semantic scene features, we enable skill generalization across diverse object poses and spatial configurations. Our physical robot experiments validate that our approach preserves the structural constraints of the task while adapting to challenging environment variations. These results highlight the promise of geometry-driven, semantically grounded reference frames as a foundation for scalable imitation learning. Moreover, our framework is agnostic to the type of downstream motion primitives; future works therefore could explore beyond DMPs such as probabilistic movement primitives and kernel-based models. Future works can also extend our approach with richer object representations to enable grasping strategies across diverse geometries.
\section{Limitations}
Since \textbf{TReF-6} relies on Grounded-SAM for localization, semantic mis-segmentation occasionally occurs. Since segmentation is not fine-tuned and operates out-of-the-box, addressing such external semantic ambiguities remains future work.

\textbf{TReF-6} extracts a single influence point to define a task-relevant 6DoF frame. While more complex spatial constraints could theoretically require richer representations, our experiments demonstrate that a single frame abstraction is sufficient to generalize across diverse atomic tasks in practice.

\textbf{TReF-6} focuses on motion generation after an object has been acquired, modeling the tool or object as a single representative point. It does not address grasp planning or complex contact interactions, and thus assumes that a feasible grasp has already been achieved prior to motion execution. Despite this simplification, experiments demonstrate that \textbf{TReF-6} achieved near-perfect success rates in task execution once the object is properly grasped. Future work will focus on extending the framework to integrate grasp planning, enabling the system to jointly reason about how to grasp and how to perform the task within the same spatial frame.
\section{Acknowledgments}

The authors would like to thank members of the Yale Inquisitive Robotics Lab and Qian Wang for their feedback and contributions.

\clearpage


\bibliography{output}  

\begin{thebibliography}{41}
\providecommand{\natexlab}[1]{#1}
\providecommand{\url}[1]{\texttt{#1}}
\expandafter\ifx\csname urlstyle\endcsname\relax
  \providecommand{\doi}[1]{doi: #1}\else
  \providecommand{\doi}{doi: \begingroup \urlstyle{rm}\Url}\fi

\bibitem[Intelligence et~al.(2025)Intelligence, Black, Brown, Darpinian, Dhabalia, Driess, Esmail, Equi, Finn, Fusai, Galliker, Ghosh, Groom, Hausman, Ichter, Jakubczak, Jones, Ke, LeBlanc, Levine, Li-Bell, Mothukuri, Nair, Pertsch, Ren, Shi, Smith, Springenberg, Stachowicz, Tanner, Vuong, Walke, Walling, Wang, Yu, and Zhilinsky]{intelligence2025pi05visionlanguageactionmodelopenworld}
P.~Intelligence, K.~Black, N.~Brown, J.~Darpinian, K.~Dhabalia, D.~Driess, A.~Esmail, M.~Equi, C.~Finn, N.~Fusai, M.~Y. Galliker, D.~Ghosh, L.~Groom, K.~Hausman, B.~Ichter, S.~Jakubczak, T.~Jones, L.~Ke, D.~LeBlanc, S.~Levine, A.~Li-Bell, M.~Mothukuri, S.~Nair, K.~Pertsch, A.~Z. Ren, L.~X. Shi, L.~Smith, J.~T. Springenberg, K.~Stachowicz, J.~Tanner, Q.~Vuong, H.~Walke, A.~Walling, H.~Wang, L.~Yu, and U.~Zhilinsky.
\newblock $\pi_{0.5}$: a vision-language-action model with open-world generalization, 2025.
\newblock URL \url{https://arxiv.org/abs/2504.16054}.

\bibitem[Kleer et~al.(2023)Kleer, Rekrut, Wolter, Schwartz, and Feld]{warehouseKleerHRI}
N.~Kleer, M.~Rekrut, J.~Wolter, T.~Schwartz, and M.~Feld.
\newblock A multimodal teach-in approach to the pick-and-place problem in human-robot collaboration.
\newblock In \emph{Companion of the 2023 ACM/IEEE International Conference on Human-Robot Interaction}, HRI '23, page 81–85, New York, NY, USA, 2023. Association for Computing Machinery.
\newblock ISBN 9781450399708.
\newblock \doi{10.1145/3568294.3580047}.
\newblock URL \url{https://doi.org/10.1145/3568294.3580047}.

\bibitem[Shi et~al.(2023)Shi, Xu, Clarke, Li, and Wu]{shi2023robocook}
H.~Shi, H.~Xu, S.~Clarke, Y.~Li, and J.~Wu.
\newblock Robocook: Long-horizon elasto-plastic object manipulation with diverse tools.
\newblock \emph{arXiv preprint arXiv:2306.14447}, 2023.

\bibitem[Wen et~al.(2024)Wen, Yang, Kautz, and Birchfield]{wen2024foundationposeunified6dpose}
B.~Wen, W.~Yang, J.~Kautz, and S.~Birchfield.
\newblock Foundationpose: Unified 6d pose estimation and tracking of novel objects, 2024.
\newblock URL \url{https://arxiv.org/abs/2312.08344}.

\bibitem[Manuelli et~al.(2019)Manuelli, Gao, Florence, and Tedrake]{manuelli2019kpamkeypointaffordancescategorylevel}
L.~Manuelli, W.~Gao, P.~Florence, and R.~Tedrake.
\newblock kpam: Keypoint affordances for category-level robotic manipulation, 2019.
\newblock URL \url{https://arxiv.org/abs/1903.06684}.

\bibitem[Liu et~al.(2024)Liu, Fang, Abbeel, and Levine]{liu2024mokaopenworldroboticmanipulation}
F.~Liu, K.~Fang, P.~Abbeel, and S.~Levine.
\newblock Moka: Open-world robotic manipulation through mark-based visual prompting, 2024.
\newblock URL \url{https://arxiv.org/abs/2403.03174}.

\bibitem[Huang et~al.(2024)Huang, Wang, Li, Zhang, and Fei-Fei]{huang2024rekepspatiotemporalreasoningrelational}
W.~Huang, C.~Wang, Y.~Li, R.~Zhang, and L.~Fei-Fei.
\newblock Rekep: Spatio-temporal reasoning of relational keypoint constraints for robotic manipulation, 2024.
\newblock URL \url{https://arxiv.org/abs/2409.01652}.

\bibitem[Li et~al.(2025)Li, Sun, Aditya, and Figueroa]{li2025elasticmotionpolicyadaptive}
T.~Li, S.~Sun, S.~S. Aditya, and N.~Figueroa.
\newblock Elastic motion policy: An adaptive dynamical system for robust and efficient one-shot imitation learning, 2025.
\newblock URL \url{https://arxiv.org/abs/2503.08029}.

\bibitem[Pan et~al.(2022)Pan, Okorn, Zhang, Eisner, and Held]{pan2022taxpose}
C.~Pan, B.~Okorn, H.~Zhang, B.~Eisner, and D.~Held.
\newblock {TAX}-pose: Task-specific cross-pose estimation for robot manipulation.
\newblock In \emph{6th Annual Conference on Robot Learning}, 2022.
\newblock URL \url{https://openreview.net/forum?id=YmJi0bTfeNX}.

\bibitem[Simeonov et~al.(2022)Simeonov, Du, Tagliasacchi, Tenenbaum, Rodriguez, Agrawal, and Sitzmann]{se3representation2022}
A.~Simeonov, Y.~Du, A.~Tagliasacchi, J.~B. Tenenbaum, A.~Rodriguez, P.~Agrawal, and V.~Sitzmann.
\newblock Neural descriptor fields: Se(3)-equivariant object representations for manipulation.
\newblock In \emph{2022 International Conference on Robotics and Automation (ICRA)}, pages 6394--6400, 2022.
\newblock \doi{10.1109/ICRA46639.2022.9812146}.

\bibitem[Ochoa et~al.(2024)Ochoa, Oh, Kwon, Domae, and Matsubara]{subgoalOchoa2024}
C.~Ochoa, H.~Oh, Y.~Kwon, Y.~Domae, and T.~Matsubara.
\newblock Ispil: Interactive sub-goal-planning imitation learning for long-horizon tasks with diverse goals.
\newblock \emph{IEEE Access}, 12:\penalty0 197616--197631, 2024.
\newblock \doi{10.1109/ACCESS.2024.3521302}.

\bibitem[Chang et~al.(2024)Chang, Boularias, and Jain]{oneShotVisualContactRich}
H.~Chang, A.~Boularias, and S.~Jain.
\newblock Insert-one: One-shot robust visual-force servoing for novel object insertion with 6-dof tracking.
\newblock In \emph{2024 IEEE/RSJ International Conference on Intelligent Robots and Systems (IROS)}, pages 2935--2942, 2024.
\newblock \doi{10.1109/IROS58592.2024.10801884}.

\bibitem[Winn et~al.(2012)Winn, Gao, Mishra, and Julius]{obstacleWinn2012}
A.~Winn, X.~Gao, S.~Mishra, and A.~A. Julius.
\newblock Learning potential functions by demonstration for path planning.
\newblock In \emph{2012 IEEE 51st IEEE Conference on Decision and Control (CDC)}, pages 4654--4659, 2012.
\newblock \doi{10.1109/CDC.2012.6426153}.

\bibitem[Subramani et~al.(2018)Subramani, Zinn, and Gleicher]{constraintGuru2018}
G.~Subramani, M.~Zinn, and M.~Gleicher.
\newblock Inferring geometric constraints in human demonstrations.
\newblock In A.~Billard, A.~Dragan, J.~Peters, and J.~Morimoto, editors, \emph{Proceedings of The 2nd Conference on Robot Learning}, volume~87 of \emph{Proceedings of Machine Learning Research}, pages 223--236. PMLR, 29--31 Oct 2018.
\newblock URL \url{https://proceedings.mlr.press/v87/subramani18a.html}.

\bibitem[Bestick et~al.(2018)Bestick, Pandya, Bajcsy, and Dragan]{humanPreferenceBestick2018}
A.~Bestick, R.~Pandya, R.~Bajcsy, and A.~D. Dragan.
\newblock Learning human ergonomic preferences for handovers.
\newblock In \emph{2018 IEEE International Conference on Robotics and Automation (ICRA)}, pages 3257--3264, 2018.
\newblock \doi{10.1109/ICRA.2018.8461216}.

\bibitem[Jin et~al.(2020)Jin, Petrich, Zhang, Dehghan, and Jagersand]{geometricJun2020}
J.~Jin, L.~Petrich, Z.~Zhang, M.~Dehghan, and M.~Jagersand.
\newblock Visual geometric skill inference by watching human demonstration.
\newblock In \emph{2020 IEEE International Conference on Robotics and Automation (ICRA)}, page 8985–8991. IEEE, May 2020.
\newblock \doi{10.1109/icra40945.2020.9196570}.
\newblock URL \url{http://dx.doi.org/10.1109/ICRA40945.2020.9196570}.

\bibitem[Zhang and Lee(2025)]{zhang2025iaaointeractiveaffordancelearning}
C.~Zhang and G.~H. Lee.
\newblock Iaao: Interactive affordance learning for articulated objects in 3d environments, 2025.
\newblock URL \url{https://arxiv.org/abs/2504.06827}.

\bibitem[Ijspeert et~al.(2002)Ijspeert, Nakanishi, and Schaal]{ijspeert2002movement}
A.~Ijspeert, J.~Nakanishi, and S.~Schaal.
\newblock Movement imitation with nonlinear dynamical systems in humanoid robots.
\newblock In \emph{Proceedings 2002 IEEE International Conference on Robotics and Automation (Cat. No.02CH37292)}, volume~2, pages 1398--1403 vol.2, 2002.
\newblock \doi{10.1109/ROBOT.2002.1014739}.

\bibitem[Ijspeert et~al.(2013)Ijspeert, Nakanishi, Hoffmann, Pastor, and Schaal]{ijspeert2013dynamical}
A.~J. Ijspeert, J.~Nakanishi, H.~Hoffmann, P.~Pastor, and S.~Schaal.
\newblock Dynamical movement primitives: Learning attractor models for motor behaviors.
\newblock \emph{Neural Computation}, 25\penalty0 (2):\penalty0 328--373, 2013.
\newblock \doi{10.1162/NECO_a_00393}.

\bibitem[Pastor et~al.(2009)Pastor, Hoffmann, Asfour, and Schaal]{pastor2009learning}
P.~Pastor, H.~Hoffmann, T.~Asfour, and S.~Schaal.
\newblock Learning and generalization of motor skills by learning from demonstration.
\newblock In \emph{2009 IEEE International Conference on Robotics and Automation}, pages 763--768, 2009.
\newblock \doi{10.1109/ROBOT.2009.5152385}.

\bibitem[Koutras and Doulgeri(2020)]{frameDMP}
L.~Koutras and Z.~Doulgeri.
\newblock A novel dmp formulation for global and frame independent spatial scaling in the task space.
\newblock In \emph{2020 29th IEEE International Conference on Robot and Human Interactive Communication (RO-MAN)}, pages 727--732, 2020.
\newblock \doi{10.1109/RO-MAN47096.2020.9223500}.

\bibitem[Calinon(2016)]{Calinon16JIST}
S.~Calinon.
\newblock A tutorial on task-parameterized movement learning and retrieval.
\newblock \emph{Intelligent Service Robotics}, 9\penalty0 (1):\penalty0 1--29, January 2016.
\newblock ISSN 1861-2776.
\newblock \doi{10.1007/s11370-015-0187-9}.

\bibitem[Huang et~al.(2018)Huang, Silvério, Rozo, and Caldwell]{huang2018generalizedtaskparameterizedskilllearning}
Y.~Huang, J.~Silvério, L.~Rozo, and D.~G. Caldwell.
\newblock Generalized task-parameterized skill learning, 2018.
\newblock URL \url{https://arxiv.org/abs/1707.01696}.

\bibitem[Zhu et~al.(2022)Zhu, Gienger, and Kober]{zhu2022learningtaskparameterizedskillsdemonstrations}
J.~Zhu, M.~Gienger, and J.~Kober.
\newblock Learning task-parameterized skills from few demonstrations, 2022.
\newblock URL \url{https://arxiv.org/abs/2201.09975}.

\bibitem[Hu and Kuchenbecker(2019)]{hu2019hierarchical}
S.~Hu and K.~J. Kuchenbecker.
\newblock Hierarchical task-parameterized learning from demonstration for collaborative object movement.
\newblock \emph{Applied Bionics and Biomechanics}, 2019:\penalty0 9765383, 2019.
\newblock \doi{10.1155/2019/9765383}.

\bibitem[Gibson(2014)]{gibson2014ecological}
J.~J. Gibson.
\newblock \emph{The Ecological Approach to Visual Perception: Classic Edition}.
\newblock Psychology Press, 1st edition, 2014.
\newblock \doi{10.4324/9781315740218}.
\newblock URL \url{https://doi.org/10.4324/9781315740218}.

\bibitem[Yen-Chen et~al.(2022)Yen-Chen, Florence, Zeng, Barron, Du, Ma, Simeonov, Garcia, and Isola]{yenchen2022miramentalimageryrobotic}
L.~Yen-Chen, P.~Florence, A.~Zeng, J.~T. Barron, Y.~Du, W.-C. Ma, A.~Simeonov, A.~R. Garcia, and P.~Isola.
\newblock Mira: Mental imagery for robotic affordances, 2022.
\newblock URL \url{https://arxiv.org/abs/2212.06088}.

\bibitem[Bahl et~al.(2023)Bahl, Mendonca, Chen, Jain, and Pathak]{bahl2023affordanceshumanvideosversatile}
S.~Bahl, R.~Mendonca, L.~Chen, U.~Jain, and D.~Pathak.
\newblock Affordances from human videos as a versatile representation for robotics, 2023.
\newblock URL \url{https://arxiv.org/abs/2304.08488}.

\bibitem[Chen et~al.(2023)Chen, Gao, Lin, and Shou]{chen2023affordancegroundingdemonstrationvideo}
J.~Chen, D.~Gao, K.~Q. Lin, and M.~Z. Shou.
\newblock Affordance grounding from demonstration video to target image, 2023.
\newblock URL \url{https://arxiv.org/abs/2303.14644}.

\bibitem[Cheng et~al.(2023)Cheng, Wu, Shen, Ning, Zhan, and Dong]{cheng2023learningenvironmentawareaffordance3d}
K.~Cheng, R.~Wu, Y.~Shen, C.~Ning, G.~Zhan, and H.~Dong.
\newblock Learning environment-aware affordance for 3d articulated object manipulation under occlusions, 2023.
\newblock URL \url{https://arxiv.org/abs/2309.07510}.

\bibitem[Jang et~al.(2022)Jang, Irpan, Khansari, Kappler, Ebert, Lynch, Levine, and Finn]{BC-Z}
E.~Jang, A.~Irpan, M.~Khansari, D.~Kappler, F.~Ebert, C.~Lynch, S.~Levine, and C.~Finn.
\newblock Bc-z: Zero-shot task generalization with robotic imitation learning.
\newblock In A.~Faust, D.~Hsu, and G.~Neumann, editors, \emph{Proceedings of the 5th Conference on Robot Learning}, volume 164 of \emph{Proceedings of Machine Learning Research}, pages 991--1002. PMLR, 08--11 Nov 2022.
\newblock URL \url{https://proceedings.mlr.press/v164/jang22a.html}.

\bibitem[Hadjivelichkov et~al.(2023)Hadjivelichkov, Zwane, Deisenroth, Agapito, and Kanoulas]{Hadjivelichkov2022affcorrs}
D.~Hadjivelichkov, S.~Zwane, M.~Deisenroth, L.~Agapito, and D.~Kanoulas.
\newblock {One-Shot Transfer of Affordance Regions? AffCorrs!}
\newblock In K.~Liu, D.~Kulic, and J.~Ichnowski, editors, \emph{{Proceedings of The 6th Conference on Robot Learning (CoRL)}}, volume 205 of \emph{Proceedings of Machine Learning Research}, pages 550--560, 14--18 Dec 2023.

\bibitem[Vitiello et~al.(2023)Vitiello, Dreczkowski, and Johns]{vitiello2023oneshotimitationlearningpose}
P.~Vitiello, K.~Dreczkowski, and E.~Johns.
\newblock One-shot imitation learning: A pose estimation perspective, 2023.
\newblock URL \url{https://arxiv.org/abs/2310.12077}.

\bibitem[Zhang and Boularias(2024)]{InvariantRegionIn3D}
X.~Zhang and A.~Boularias.
\newblock One-shot imitation learning with invariance matching for robotic manipulation.
\newblock In \emph{Proceedings of the Robotics: Science and Systems (RSS)}, 07 2024.
\newblock \doi{10.15607/RSS.2024.XX.134}.

\bibitem[Yin and Abbeel(2024)]{yin2024offline}
Z.-H. Yin and P.~Abbeel.
\newblock Offline imitation learning through graph search and retrieval.
\newblock \emph{Robotics: Science and Systems}, 2024.

\bibitem[Ajay et~al.(2023)Ajay, Han, Du, Li, Gupta, Jaakkola, Tenenbaum, Kaelbling, Srivastava, and Agrawal]{ajay2023compositionalfoundationmodelshierarchical}
A.~Ajay, S.~Han, Y.~Du, S.~Li, A.~Gupta, T.~Jaakkola, J.~Tenenbaum, L.~Kaelbling, A.~Srivastava, and P.~Agrawal.
\newblock Compositional foundation models for hierarchical planning, 2023.
\newblock URL \url{https://arxiv.org/abs/2309.08587}.

\bibitem[Gopinath and Argall(2020)]{9066939}
D.~E. Gopinath and B.~D. Argall.
\newblock Active intent disambiguation for shared control robots.
\newblock \emph{IEEE Transactions on Neural Systems and Rehabilitation Engineering}, 28\penalty0 (6):\penalty0 1497--1506, 2020.
\newblock \doi{10.1109/TNSRE.2020.2987878}.

\bibitem[Candès et~al.(2015)Candès, Li, and Soltanolkotabi]{Cand_s_2015}
E.~J. Candès, X.~Li, and M.~Soltanolkotabi.
\newblock Phase retrieval via wirtinger flow: Theory and algorithms.
\newblock \emph{IEEE Transactions on Information Theory}, 61\penalty0 (4):\penalty0 1985–2007, Apr. 2015.
\newblock ISSN 1557-9654.
\newblock \doi{10.1109/tit.2015.2399924}.
\newblock URL \url{http://dx.doi.org/10.1109/TIT.2015.2399924}.

\bibitem[Chi et~al.(2019)Chi, Lu, and Chen]{Chi_2019}
Y.~Chi, Y.~M. Lu, and Y.~Chen.
\newblock Nonconvex optimization meets low-rank matrix factorization: An overview.
\newblock \emph{IEEE Transactions on Signal Processing}, 67\penalty0 (20):\penalty0 5239–5269, Oct. 2019.
\newblock ISSN 1941-0476.
\newblock \doi{10.1109/tsp.2019.2937282}.
\newblock URL \url{http://dx.doi.org/10.1109/TSP.2019.2937282}.

\bibitem[Kingma and Ba(2014)]{kingma2014adam}
D.~P. Kingma and J.~Ba.
\newblock Adam: A method for stochastic optimization.
\newblock \emph{arXiv preprint arXiv:1412.6980}, 2014.

\bibitem[OpenAI(2024)]{openai2024gpt4o}
OpenAI.
\newblock Hello gpt-4o, 2024.
\newblock URL \url{https://openai.com/index/hello-gpt-4o/}.

\end{thebibliography}

\clearpage
\section{Appendix}

\subsection{Loss Landscape}
\begin{figure}[h]
    \centering
    \includegraphics[width=0.75\linewidth]{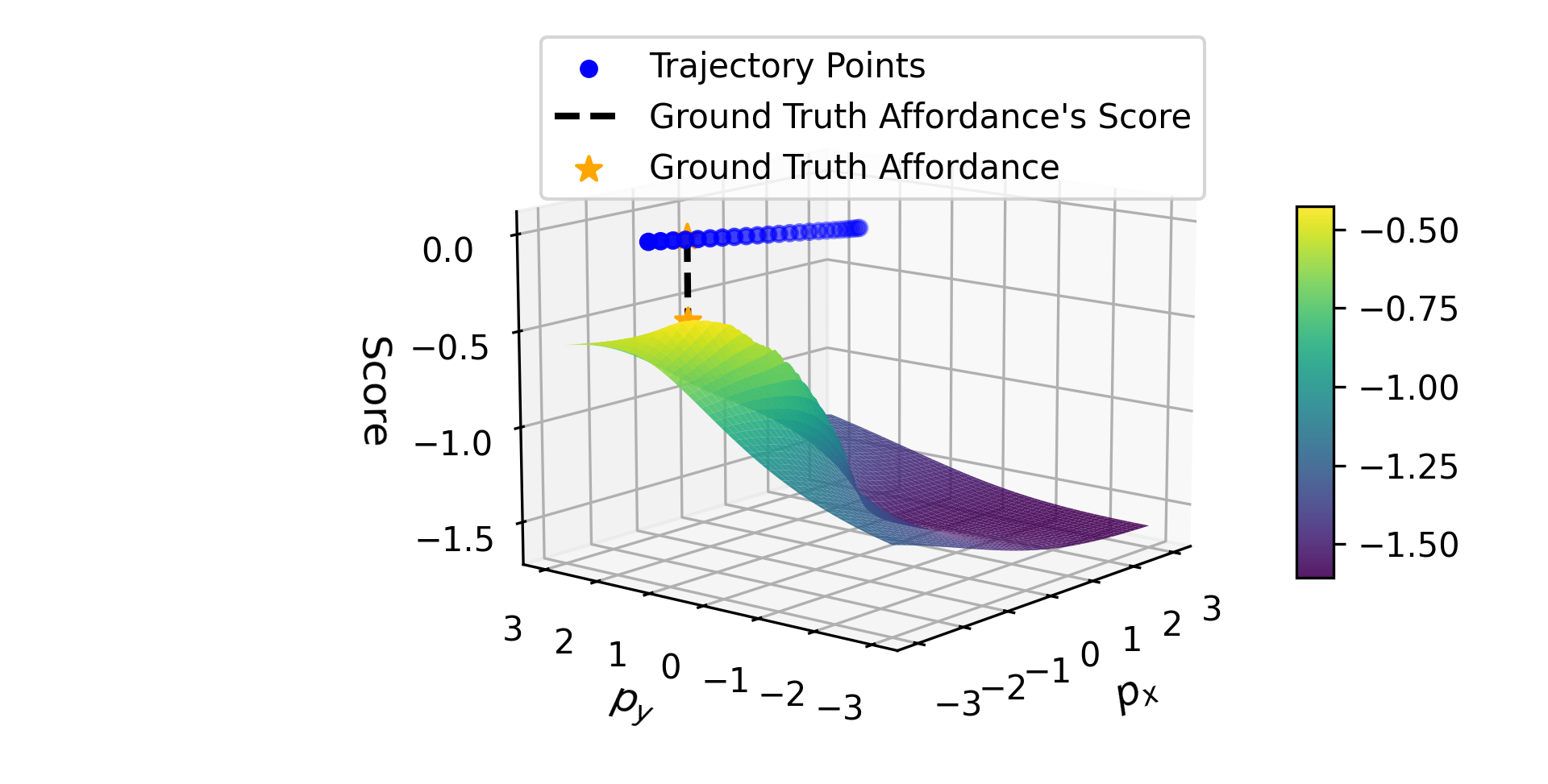}
    \caption{Score Landscape in 2D case for a trajectory length of $T = 25$. Notice the large flat gradient areas around regions far from the trajectory.}
    \label{fig:non-convex}
\end{figure}

\FloatBarrier
\label{Convex}
\subsection{Definition of Baselines}
\label{app:surrogate}

\textbf{A. Cosine Similarity Score}

We define the \textbf{cosine similarity score} as:
\begin{equation}
    \text{Score}_{\text{cos}}(p) = \frac{1}{T}\sum^T_{t=1}\left| \frac{(p - x_t)^\top \Ddot{x}_t}{||(p - x_t)|| \cdot ||\Ddot{x}_t|| + \epsilon}\right|
\end{equation}
While still \textbf{non-convex}, cosine similarity score is effective in 2D due to a constrained directional space and sharp gradient alignment. Its performance degrades in 3D settings where directional ambiguity, vanishing gradients, and the lack of magnitude information significantly impair its ability to infer correct influence points.

\textbf{B. Quadratic Residual Score}

We define the \textbf{quadratic residual score} as:
\begin{equation}
    \text{Score}_{\text{quad}}(p) = \frac{1}{T}\sum^T_{t=1}|| (p-x_t) - \Ddot{x}_t||^2
\end{equation}
Despite this objective being \textbf{convex} and easy to optimize, it assumes the magnitude of the observed acceleration matches that of the position-based prediction, which is often violated in realistic demonstrations and results in low accuracy even in simulated environment. 

\textbf{C. Inverse Dynamics Triangulation}

Under the Newtonian assumption $\vec{f}_t = m\cdot \vec{a}_t$, the \textbf{inverse dynamics triangulation} method interpret the direction of the acceleration vector at each timestep as a ray pointing toward the latent affordance point. Then, the triangulation process estimates the point in space that minimizes the orthogonal distance to all such rays:

\begin{equation}
    \min_{p \in \mathbb{R}^3} \sum_{t=1}^T \left| (I - \hat{a}_t \hat{a}_t^\top)(x_t - p) \right|^2
\end{equation}

where $\hat{a}_t$ is the unit acceleration direction at timestep $t$, and $(I -\hat{a}_t \hat{a}_t^\top)$ is the projection matrix orthogonal to that direction. 

While effective under clean conditions, this method is highly sensitive to noise in the direction of $\vec{a}_t$, especially in higher dimensions where ray intersections are less geometrically constrained. As such, it performs well at low noise levels but deteriorates rapidly when acceleration signals are noisy or inconsistent.

\FloatBarrier
\subsection{Demonstrations for Each Task}
\begin{figure}[h]
    \centering
    \includegraphics[width=0.95\textwidth, trim={0 200 0 0}, clip]{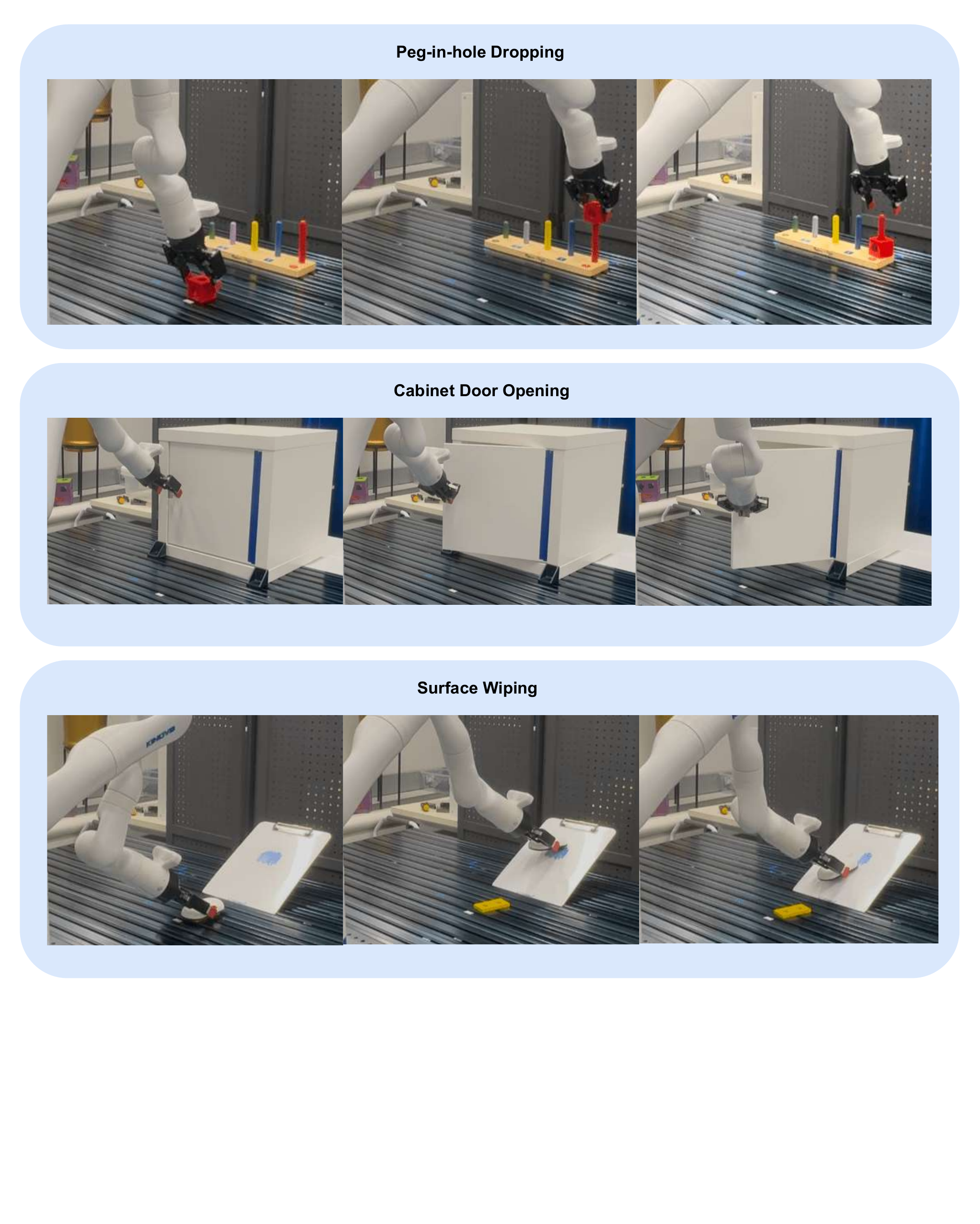}
    \caption{\textbf{Single Demonstration per Task.} Each row shows the provided demonstration for one of the three tasks: peg-in-hole dropping (top), cabinet door opening (middle), and surface wiping (bottom). These single demonstrations are the only inputs used for learning; our method infers task-relevant spatial reference frames from each to enable downstream generalization to novel object configurations and orientations.}
    \label{fig:demonstration}
\end{figure}

\FloatBarrier
\subsection{Simulation Experiment Details}
\label{sec:simulation_more}
\subsubsection{3D Baseline Analysis}

\begin{table*}[t!]
\caption{Comparison of Spatial influence Inference Methods under Different Noise Levels over 50 Random Seeds in 3D. Report in MEDE: Mean $\pm$ Std}
\label{tab:influence_comparison}
\centering
\small
\resizebox{\textwidth}{!}{
\begin{tabular}{lccccc}
\toprule
\textbf{Method} & \textbf{0\% Noise} & \textbf{10\% Noise} & \textbf{30\% Noise} & \textbf{50\% Noise} & \textbf{80\% Noise} \\
\midrule
Directional Consistency Score   & $0.0263 \pm 0.0392$ & $\mathbf{0.0560 \pm 0.0276}$ & $\mathbf{0.1688 \pm 0.0831}$ & $\mathbf{0.3044 \pm 0.1808}$ & $\mathbf{0.5814 \pm 0.3722}$ \\
Directional Consistency Score (Random Initialization) & $0.1884 \pm 0.8217$ & $0.2794 \pm 1.0735$ & $0.3076 \pm 0.6416$ & $0.4828 \pm 1.1278$ & $0.7945 \pm 1.2573$ \\
Inverse Dynamics Triangulation & $\mathbf{0.0127 \pm 0.0157}$ & $0.1111 \pm 0.0818$ & $0.4009 \pm 0.2589$ & $0.6540 \pm 0.4020$ & $0.9474 \pm 0.5689$ \\
Cosine Similarity & $1.8483 \pm 2.2455$ & $1.8892 \pm 2.3067$ & $1.4667 \pm 2.0082$ & $1.1981 \pm 1.7585$ & $1.7815 \pm 2.0691$ \\
Quadratic Residual & $0.3311 \pm 0.2217$ & $0.3397 \pm 0.2225$ & $0.4147 \pm 0.2441$ & $0.5191 \pm 0.2868$ & $0.7482 \pm 0.4037$ \\
\bottomrule
\end{tabular}
}
\end{table*}

\textbf{Directional Consistency Score (ours)} achieved the best overall performance across all noise levels tested. Even as the force signal became increasingly corrupted, it maintained both low error and low variance: MEDE raises modestly from $0.0560 \pm 0.0276$ at $10\%$ noise to $0.5814 \pm 0.3722$ at $80\%$ noise. Despite its non-convexity, our method yielded stable and accurate estimates when paired with structured initialization and optimization. 

\textbf{Inverse Dynamics Triangulation} performed well at low noise levels: $0.1111 \pm 0.0818$ at $10\%$ and achieved the best performance over other methods at $0\%$ noise level with MEDE $0.0127 \pm 0.0157$. However, it degraded more rapidly as the noise increases, reaching $0.9747 \pm 0.5689$ at $80\%$ noise. We anticipated this behavior since this approach assumes clean Newtonian force signals and becomes unstable when acceleration vectors fluctuate or intersect imprecisely.

\textbf{Cosine Similarity Score} aligned force directions but disregarded magnitude. In 3D, where the directional ambiguity is high and the gradients flatten near $\cos(\theta) \approx 1$, the method struggled to guide the optimization effectively. Interestingly, increasing the level of noise improved its performance, and it reached its best performance at the noise level $50\%$ with the MEDE of $1.1981 \pm 1.7585$. The cause would be the injected noise that pushes the optimization out of local minima. It produced the highest variance overall and failed catastrophically at times, especially under low noise.

\textbf{Quadratic Residual Score} offered a convex alternative, minimizing the L2 difference between $p - x_i$ and $a_t$. However, this formulation implicitly assumes that the observed acceleration vectors have magnitudes that are consistent with the inferred direction vectors. In our simulation, we intentionally violated this assumption by scaling the force magnitude as the distance $||p-x_t||$ multiplied by a stochastic coefficient drawn from a uniform distribution. This design mimicked real-world conditions, where human-applied forces are irregular and not strictly distance-proportional. As a result, the quadratic score performed poorly, with a MEDE of $0.3311 \pm 0.2217$ even in the noise-free setting.

The results were shown in Appendix Table \ref{tab:influence_comparison}.
\subsubsection{Lower-Dimensional Intuition}

To better understand the behaviors of different scoring functions, we also conducted a controlled 2D version of the influence inference experiment with setup mirrored the 3D scenario. We evaluated the same four methods (Directional Consistency Score, Inverse Dynamics Triangulation, Cosine Similarity, and Quadratic Residual) across 50 seeds under different directional noise. The results were summarized in Appendix Table \ref{tab:influence_comparison_2d}.

In the 2D setting, both the \textbf{Directional Consistency Score (ours)} and \textbf{Inverse Dynamics Triagnulation} demonstrated strong performance under low to moderate noise. Notably, the \textbf{Directional Consistency Score} achieved the most robust accuracy across $10\%$, $30\%$, and $50\%$ noise levels, which confirmed its resilience to force perturbations even in lower-dimensional dynamics. The increased standard deviation of the \textbf{Directional Consistency Score} at $80\%$ noise could be attributed to degraded initialization under extreme noise, where the top-$k$ acceleration magnitudes no longer reliably reflected actual force directions. As a result, optimization was more likely to begin near outliers or flat-gradient regions, leading to wider variability in outcomes.

\textbf{Inverse Dynamics Triangulation} again achieved the lowest MEDE of $0.0128 \pm 0.0169$ in the noise-free setting. Although its performance degraded with increasing noise, the deterioration was more moderate than in the 3D case due to reduced directional ambiguity.

\textbf{Quadratic Residual Score}, as expected, remained stable due to its convex formulation. While it outperformed other methods at noise level $80\%$ with a MEDE of $0.4468 \pm 0.2454$, this performance was not significantly better than that of the \textbf{Directional Consistency Score}, which achieved a comparable MEDE of $0.4965 \pm 1.0663$.

\begin{table*}[t!]
\caption{Comparison of Spatial influence Inference Methods under Different Noise Levels over 50 Random Seeds in 2D. Report in MEDE: Mean $\pm$ Std}
\label{tab:influence_comparison_2d}
\centering
\small
\resizebox{\textwidth}{!}{
\begin{tabular}{lccccc}
\toprule
\textbf{Method} & \textbf{0\% Noise} & \textbf{10\% Noise} & \textbf{30\% Noise} & \textbf{50\% Noise} & \textbf{80\% Noise} \\
\midrule
Directional Consistency Score   & $0.0391 \pm 0.0289$ & $\mathbf{0.0584 \pm 0.0344}$ & $\mathbf{0.1473} \pm 0.1903$ & $\mathbf{0.2174 \pm 0.1725}$ & $0.4965 \pm 1.0663$ \\
Inverse Dynamics Triangulation & $\mathbf{0.0128 \pm 0.0169}$ & $0.0739 \pm 0.0642$ & $0.2817 \pm 0.2396$ & $0.4356 \pm 0.3194$ & $0.6179 \pm 0.4023$ \\
Cosine Similarity & $1.7646 \pm 2.0060$ & $1.5219 \pm 2.0972$ & $1.4158 \pm 2.2923$ & $1.6448 \pm 2.2317$ & $1.5106 \pm 1.8028$ \\
Quadratic Residual & $0.2275 \pm 0.1601$ & $0.2432 \pm 0.1475$ & $0.2753 \pm \mathbf{0.1612}$ & $0.3218 \pm 0.1919$ & $\mathbf{0.4468 \pm 0.2454}$ \\
\bottomrule
\end{tabular}
}
\end{table*}

\subsubsection{Inference from Partial Trajectories}

To assess how early the spatial intent can be reliably inferred, we evaluated our method under the constraint of partial observation. Specifically, we provided only the first $T$ timesteps of the demonstration trajectory $\{x_t, \Ddot{x}_t\}_{t=1}^T$ to the inference algorithm, and vary $T$ from $1$ to $100$. This setting emulated early recognition in human intentions when interacting with robots where only partial trajectory is observed.

\begin{figure*}[t]
    \centering
    \begin{subfigure}{0.32\linewidth}
        \includegraphics[width=\linewidth]{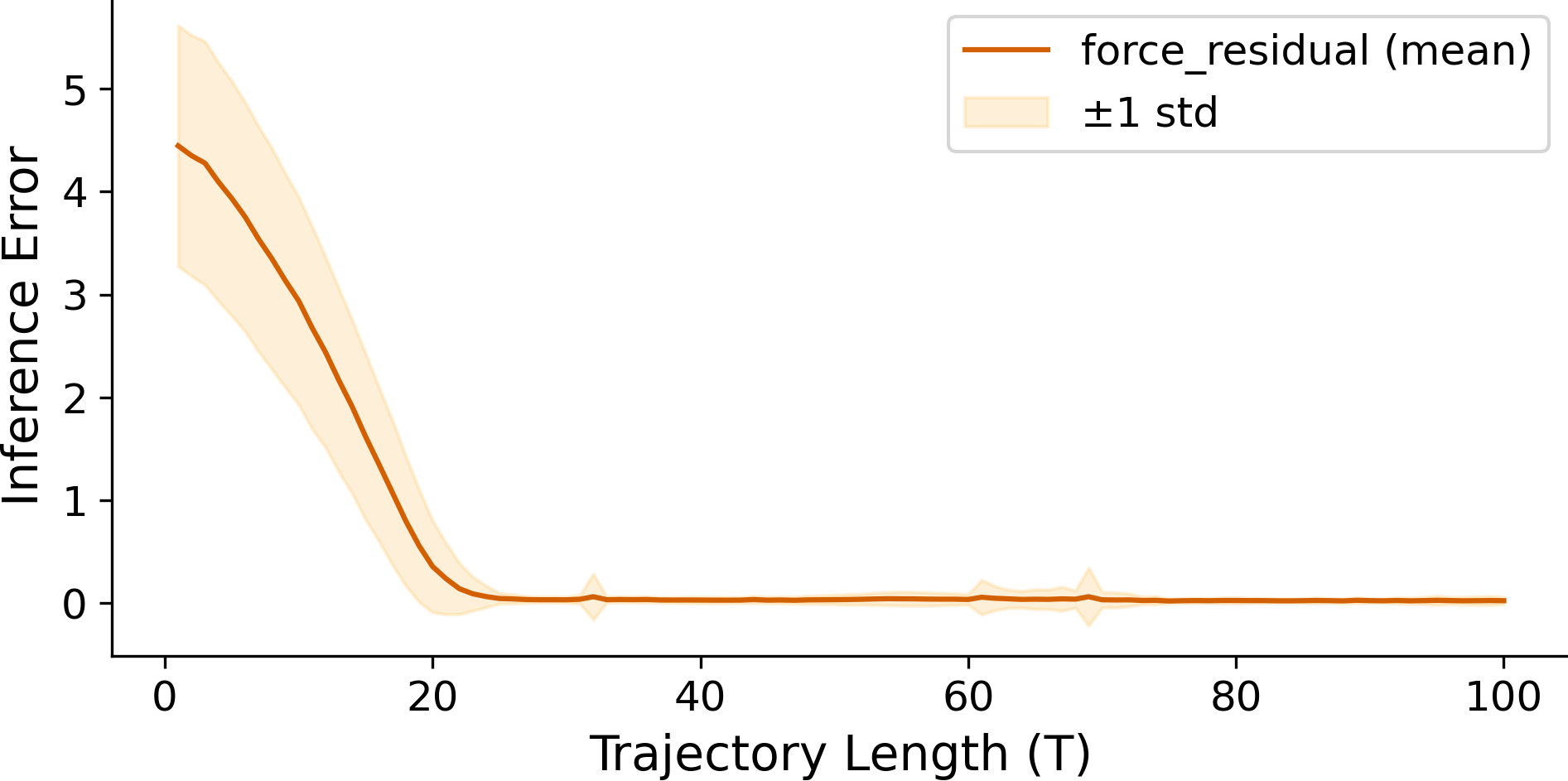}
        \caption{$0\%$ Noise}
    \end{subfigure}
    \begin{subfigure}{0.32\linewidth} 
        \includegraphics[width=\linewidth]{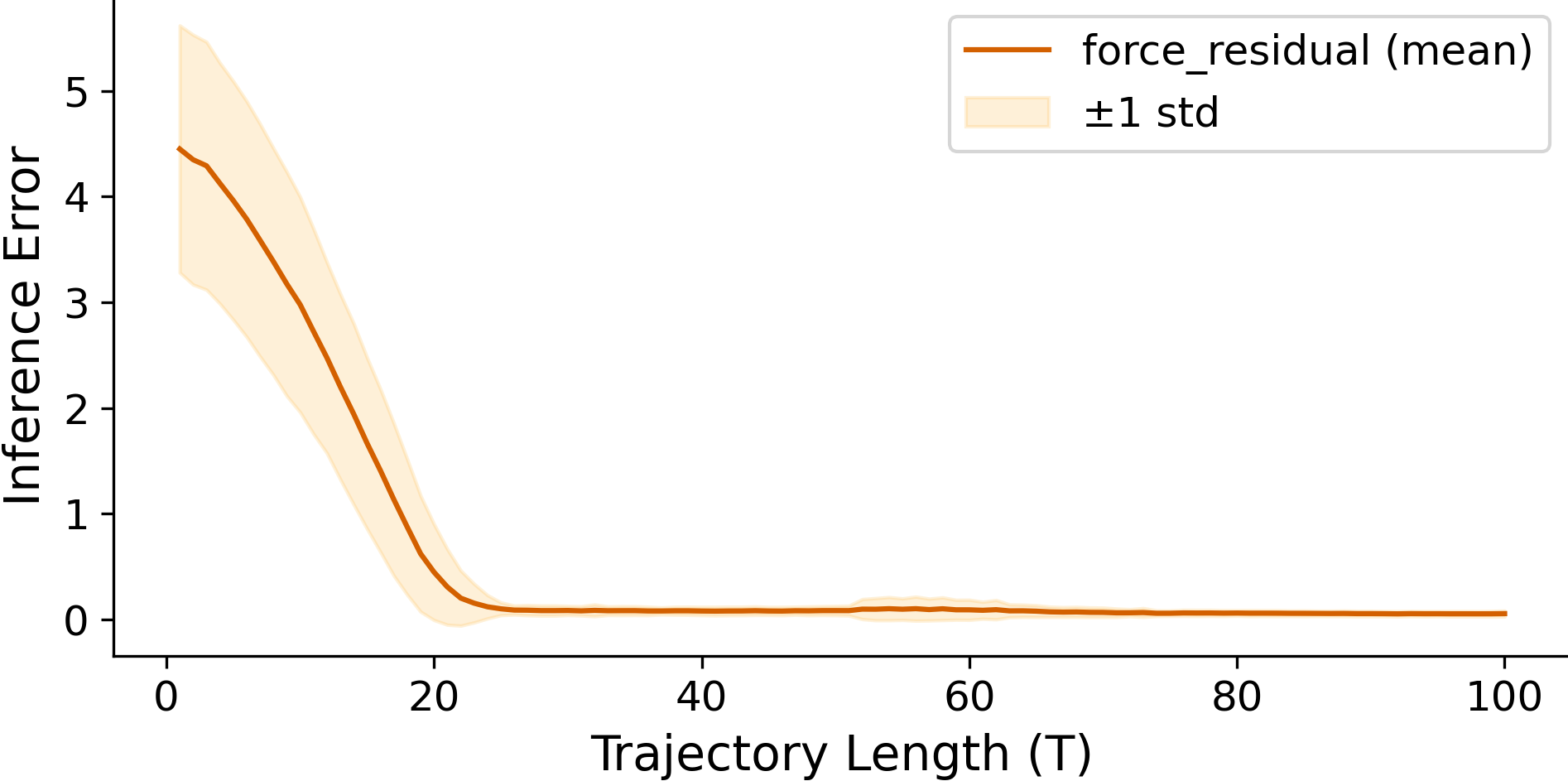} 
        \caption{$10\%$ Noise} 
    \end{subfigure} 
    \begin{subfigure}{0.32\linewidth} 
        \includegraphics[width=\linewidth]{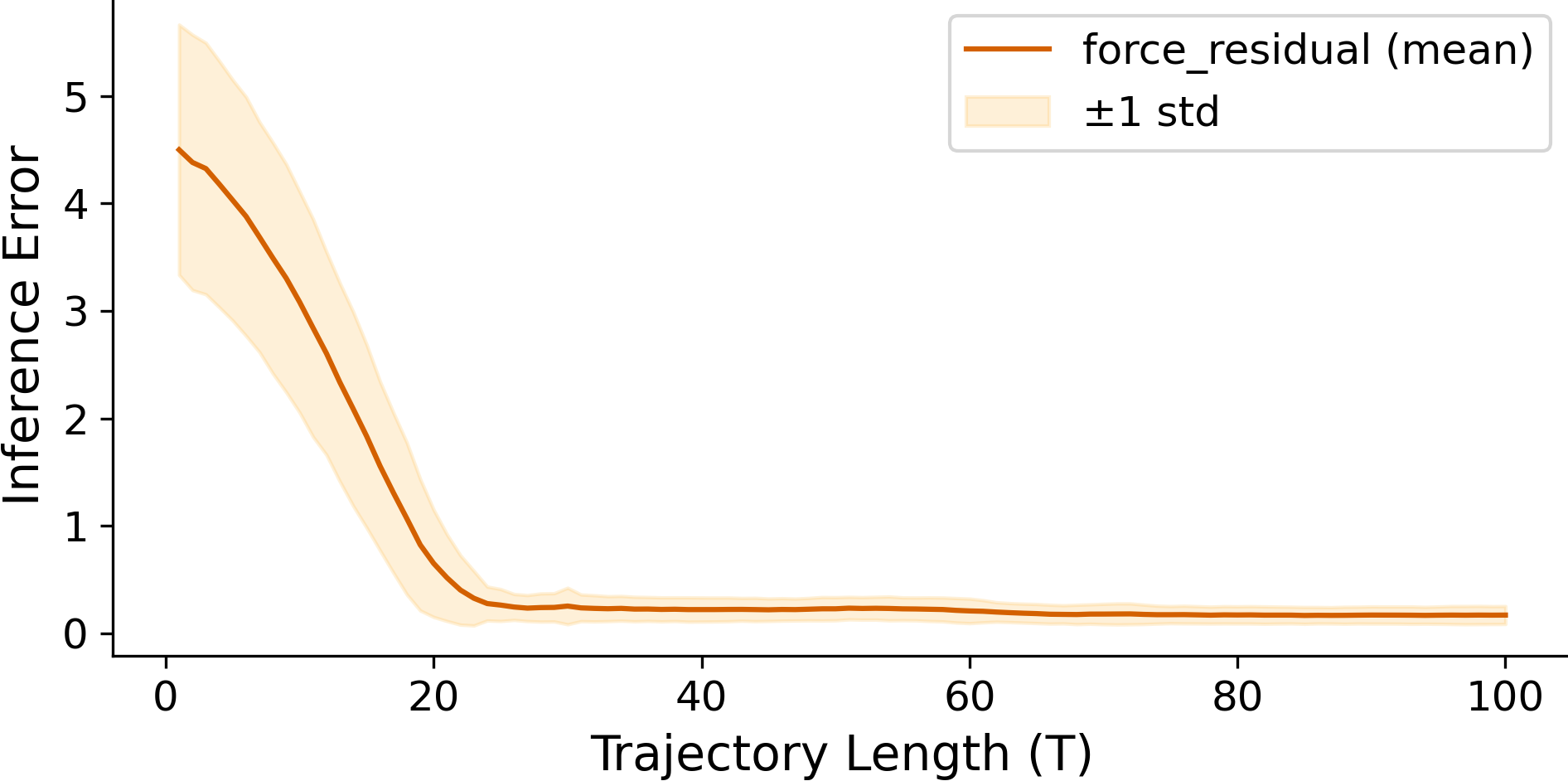} 
        \caption{$30\%$ Noise} 
    \end{subfigure} 
    \begin{subfigure}{0.32\linewidth} 
        \includegraphics[width=\linewidth]{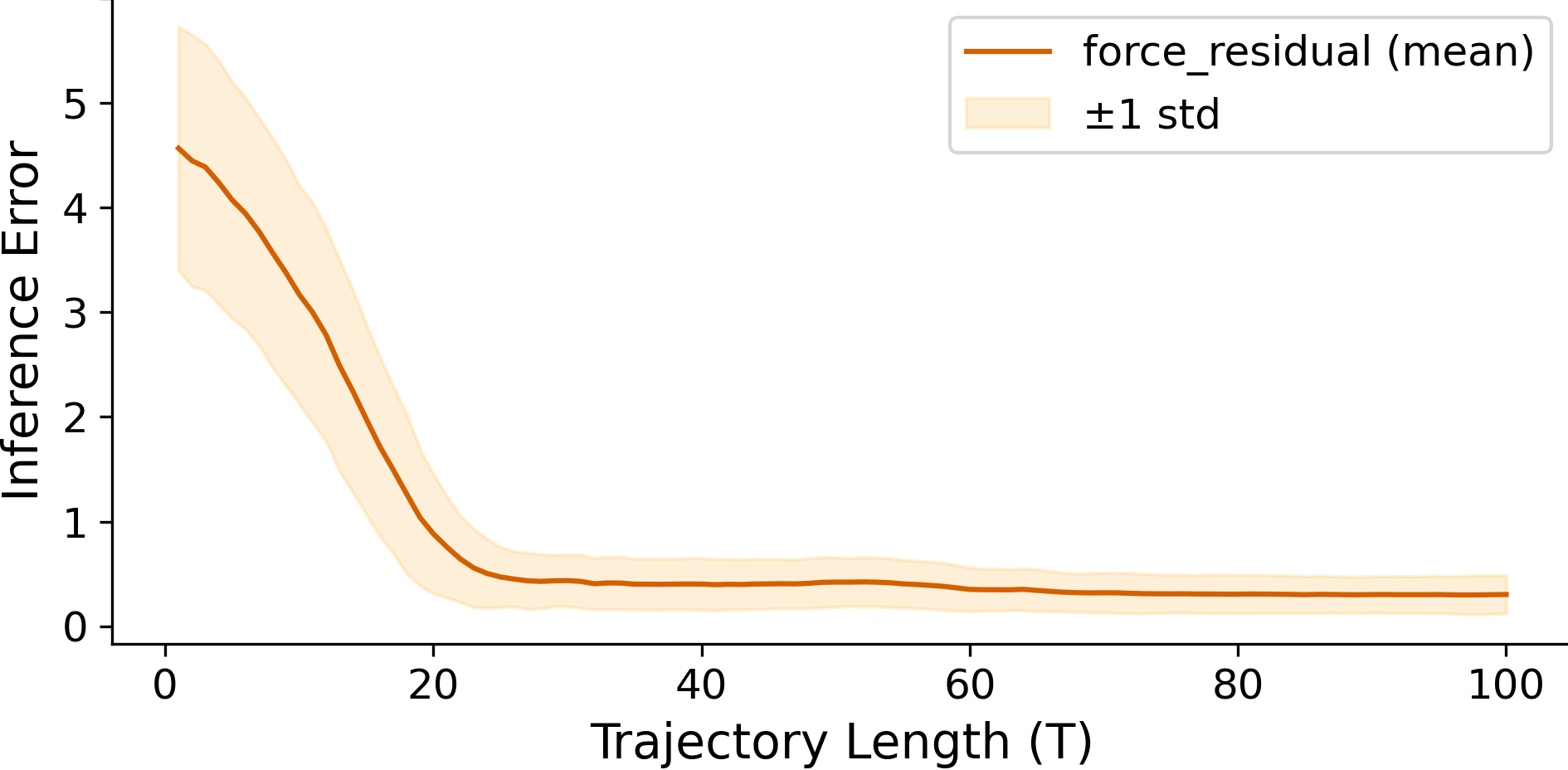} 
        \caption{$50\%$ Noise} 
    \end{subfigure} 
    \begin{subfigure}{0.32\linewidth} 
        \includegraphics[width=\linewidth]{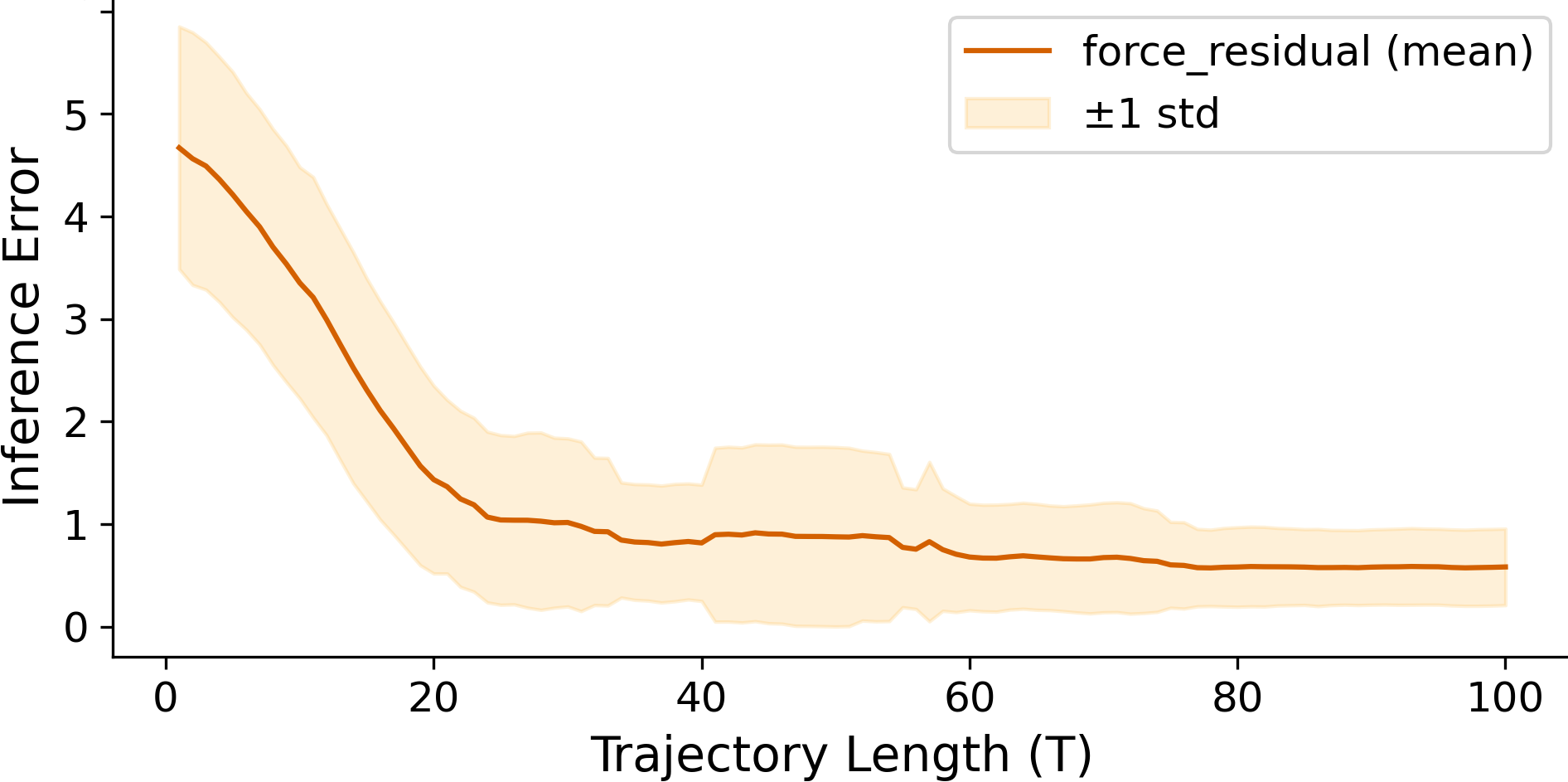} 
        \caption{$80\%$ Noise}
    \end{subfigure} 
    \caption{\textbf{Effect of Partial Trajectory on Inference Accuracy over 50 random seeds.} Mean inference error ($\pm 1$ std) of the Directional Consistency Score evaluated using the first $k$ timesteps ($\{x_t, a_t\}_{t=1}^{k}$), with $k \in [1, 100]$. Our method achieves accurate inference using less than $1/4$ of the complete trajectory under moderate noise, and remains robust even under severe corruption.}
    \label{fig:partial_traj_noise}
\end{figure*}

Appendix Figure \ref{fig:partial_traj_noise} plotted the mean inference error ($\pm 1$ std) of the fitted point across $50$ random seeds under varying trajectory lengths, for five different noise levels: $0\%$, $10\%$, $30\%$, $50\%$, and $80\%$.

Across all noise levels, we observed a significant inference error drop within the first $20 - 30$ timesteps and gradually plateaus afterward. At low noise ($0\%$, $10\%$), as few as $25$ steps sufficed to reliably identify the influence point. Under heavier noise ($50\%$, $80\%$), longer observation windows were required, but the Directional Consistency Score still converged to reasonable estimates under $40$ steps.

As expected, the variance was significantly higher when fewer steps are available, even under a noise-free setting, showing the ambiguity in intent when the motion was just beginning. As more of the trajectory was revealed, the system became increasingly accurate, suggesting that the method captured the cumulative structure of intention encoded in second-order motion.

\subsubsection{Sequential Influence Inference}

We further evaluated our method in a setting where a trajectory was governed by two spatial influence points $p_1$ and $p_2$ in sequence. The transition between the two points was defined by a known switch step, allowing the inference algorithm to segment the trajectory accordingly. For each segment, we fitted a separate influence point using our method and reported the MEDE for each inferred point.

Appendix Table \ref{tab:sequential_influence} presented the results across three noise levels ($0\%$, $10\%$, and $30\%$) and three switch ratios: $30/70$, $50/50$, and $70/30$ corresponding to the percentage of trajectory length influenced by $p_1$ vs. $p_2$. Each configuration was averaged over 50 random seeds.

Across all settings, we observed a longer influence period of a point generally yields a lower MEDE. This result aligned with intuition that longer segments provide more consistent motion patterns to constrain the optimization.

Interestingly, we found that \textbf{$30/70$ split consistently achieved the lowest overall mean error} across noise levels. This behavior could be explained by the inherent asymmetry in how motion dynamics evolved across segments. In our simulation, the first segment began with an initial velocity uniformly sampled from $[-0.5, 0.5]^3$, representing a moderate and unbiased starting momentum. In contrast, the second segment inherited the velocity from the first, which may accumulate significant directional bias by the time of switching. Consequently, estimating $p_2$ required more trajectory context.

In real-world applications such as manipulation or teleoperation, this challenge may be less significant. For instance, a human operator may pause or reorient the robot between sub-tasks, effectively resetting the velocity and making both phases more distinguishable. Our simulation represented a worst-case continuity scenario and that influence inference in practice could be even more robust given appropriately segmented demonstrations.

\begin{table*}[t]
\centering
\caption{Performance of Sequential influence Inference under Varying Noise Levels and Switch Ratios. Reported in MEDE (Mean Euclidean Distance Error) over 50 seeds.}
\label{tab:sequential_influence}
\small
\begin{tabular}{c|c|c|c|c}
\toprule
\textbf{Noise Level} & \textbf{Switch Ratio (p1 / p2)} & \textbf{MEDE (p1)} & \textbf{MEDE (p2)} & \textbf{Overall Mean Error} \\
\midrule
\multirow{3}{*}{0\%} 
    & 30 / 70 & $\mathbf{0.0551 \pm 0.0404}$ & $\mathbf{0.0541 \pm 0.0696}$ & $\mathbf{0.0546 \pm 0.0359}$ \\
    & 50 / 50 & $0.0622 \pm 0.0886$ & $0.0800 \pm 0.1255$ & $0.0711 \pm 0.0760$ \\
    & 70 / 30 & $0.0720 \pm 0.0654$ & $0.4450 \pm 0.5543$ & $0.2585 \pm 0.2733$ \\
\midrule
\multirow{3}{*}{10\%} 
    & 30 / 70 & $0.0936 \pm 0.0424$ & $\mathbf{0.0949 \pm 0.0574}$ & $\mathbf{0.0943 \pm 0.0362}$ \\
    & 50 / 50 & $0.1131 \pm 0.1572$ & $0.1337 \pm 0.1504$ & $0.1234 \pm 0.1041$ \\
    & 70 / 30 & $\mathbf{0.0888 \pm 0.0484}$ & $0.4822 \pm 0.4958$ & $0.2855 \pm 0.2459$ \\
\midrule
\multirow{3}{*}{30\%} 
    & 30 / 70 & $0.2360 \pm 0.1219$ & $\mathbf{0.2583 \pm 0.1402}$ & $\mathbf{0.2471 \pm 0.1069}$ \\
    & 50 / 50 & $0.2378 \pm 0.1643$ & $0.2799 \pm 0.2037$ & $0.2588 \pm 0.1331$ \\
    & 70 / 30 & $\mathbf{0.1955 \pm 0.1029}$ & $0.6322 \pm 0.4831$ & $0.4138 \pm 0.2442$ \\
\bottomrule
\end{tabular}
\end{table*}

\FloatBarrier
\subsection{Real World Experiment Complete Results}
\label{sec:real_world}
\begin{table}[h!]
\centering
\caption{Success Rates of Baseline DMP vs. Our Method for Peg-in-hole Dropping Task. Reported over 15 variations.}
\label{tab:success_rates_peg_in_hole}
\small
\begin{tabular}{l|c|c}
\toprule
\textbf{Metric} & \textbf{Baseline DMP} & \textbf{Ours (Gap)} \\
\midrule
Overall Success Rate & $20.0\%$ & $\mathbf{53.3\%}$ (\textcolor{darkgreen}{$\mathbf{+33.3\%}$}) \\
Grasping Success Rate & $80.0\%$ & $\mathbf{86.7\%}$ ($\mathbf{+6.7\%}$) \\
Execution Success (Given Grasp) & $25.0\%$ & $\mathbf{61.5\%}$ (\textcolor{darkgreen}{$\mathbf{+36.5\%}$}) \\
\midrule
\multicolumn{3}{l}{\textbf{Success Rate by Object:}} \\
\quad Red Cube & $1/5$ & $\mathbf{4/5}$ (\textcolor{darkgreen}{$\mathbf{+3}$}) \\
\quad Blue Cap & $1/5$ & $\mathbf{2/5}$ ($\mathbf{+1}$) \\
\quad Green Ring & $1/5$ & $\mathbf{2/5}$ ($\mathbf{+1}$) \\
\midrule
\multicolumn{3}{l}{\textbf{Success Rate by Rod:}} \\
\quad Red Rod & $\mathbf{3/3}$ & $\mathbf{3/3}$ ($+0$) \\
\quad Blue Rod & $0/3$ & $\mathbf{1/3}$ ($\mathbf{+1}$) \\
\quad Yellow Rod & $0/3$ & $\mathbf{3/3}$ (\textcolor{darkgreen}{$\mathbf{+3}$}) \\
\quad Purple Rod & $0/3$ & $0/3$ ($+0$) \\
\quad Green Rod & $0/3$ & $\mathbf{1/3}$ ($\mathbf{+1}$) \\
\bottomrule
\end{tabular}
\end{table}

\begin{table}[h!]
\centering
\caption{Success Rates of Baseline DMP vs. Our Method for Cabinet Door Opening Task. Reported over 12 variations.}
\label{tab:success_rates_door_open}
\small
\begin{tabular}{l|c|c}
\toprule
\textbf{Metric} & \textbf{Baseline DMP} & \textbf{Ours (Gap)}\\
\midrule
Overall Success Rate & $8.3\%$ & $\mathbf{66.7\%}$ (\textcolor{darkgreen}{$\mathbf{+58.3\%}$})\\
Grasping Success Rate & $16.7\%$ & $\mathbf{75.0\%}$ (\textcolor{darkgreen}{$\mathbf{+58.3\%}$})\\
Execution Success (Given Grasp) & $50.0\%$ & $\mathbf{88.9\%}$ (\textcolor{darkgreen}{$\mathbf{+38.9\%}$}) \\
\midrule
\multicolumn{3}{l}{\textbf{Success Rate by Orientation:}} \\
\quad Frontal & $1/4$ & $\mathbf{3/4}$ (\textcolor{darkgreen}{$\mathbf{+2}$})  \\
\quad Left Angled & $0/4$ & $\mathbf{3/4}$ (\textcolor{darkgreen}{$\mathbf{+3}$})  \\
\quad Right Angled & $0/4$ & $\mathbf{2/4}$ ( \textcolor{darkgreen}{$\mathbf{+2}$})\\
\midrule
\multicolumn{3}{l}{\textbf{Success Rate by Hinge Position:}} \\
\quad Hinge at Right & $1/3$ & $\mathbf{3/3}$ (\textcolor{darkgreen}{$\mathbf{+2}$}) \\
\quad Hinge at Left & $0/3$ & $\mathbf{2/3}$ (\textcolor{darkgreen}{$\mathbf{+2}$}) \\
\quad Hinge at Bottom & $0/3$ & $\mathbf{3/3}$ (\textcolor{darkgreen}{$\mathbf{+3}$}) \\
\quad Hinge at Top & $0/3$ & $0/3$ ($+0$)\\
\bottomrule
\end{tabular}
\end{table}

\begin{table}[h!]
\centering
\caption{Success Rates of Baseline DMP vs. Our Method for Surface Wiping. Reported over 9 variations.}
\label{tab:success_rates_wiping}
\small
\begin{tabular}{l|c|c}
\toprule
\textbf{Metric} & \textbf{Baseline DMP} & \textbf{Ours (Gap)} \\
\midrule
Overall Success Rate & $33.3\%$ & $\mathbf{66.7\%}$ (\textcolor{darkgreen}{$\mathbf{+33.3\%}$}) \\
Grasping Success Rate & $\mathbf{100.0\%}$ & $77.8\%$ (\textcolor{darkred}{$-22.2\%$}) \\
Execution Success (Given Grasp) & $33.3\%$ & $\mathbf{85.7\%}$ (\textcolor{darkgreen}{$\mathbf{+52.4\%}$})  \\
\midrule
\multicolumn{3}{l}{\textbf{Success Rate by Stain Color:}} \\
\quad Blue & $1/3$ & $\mathbf{2/3}$ ($\mathbf{+1}$) \\
\quad Black & $1/3$ & $\mathbf{2/3}$ ($\mathbf{+1}$) \\
\quad Red & $1/3$ & $\mathbf{2/3}$ ($\mathbf{+1}$) \\
\midrule
\multicolumn{3}{l}{\textbf{Success Rate by Surface Tilted Angle:}} \\
\quad $45^\circ$ & $\mathbf{3/3}$ & $2/3$ (\textcolor{darkred}{$-1$})  \\
\quad $90^\circ$ & $0/3$ & $\mathbf{3/3}$ (\textcolor{darkgreen}{$\mathbf{+3}$})  \\
\quad $30^\circ$ & $0/3$ & $\mathbf{1/3}$ ($\mathbf{+1}$)  \\
\bottomrule
\end{tabular}
\end{table}

\FloatBarrier
\subsection{DMP Roll Out Definition}
\label{sec:dmp}

\paragraph{Position DMP:}
\begin{equation}
\text{DMP}_{\text{pos}}(x^{\text{new}}_0) = \left\{ y_t \;\middle|\;
\begin{aligned}
\tau \ddot{y}_t &= \alpha_z \left( \beta_z (g^{\text{deploy}} - y_t) - \tau \dot{y}_t \right) + f(s_t), \\
\tau \dot{y}_t &= \ddot{y}_t, \quad \tau \dot{s}_t = -\alpha_s s_t, \\
f(s) &= \frac{\sum_i \psi_i(s) w_i}{\sum_i \psi_i(s)} s\, g^{\text{deploy}}, \quad
g^{\text{deploy}} = \Delta x_T \cdot \frac{\|p^* - x^{\text{new}}_0\|}{\|p^* - x^{\text{demo}}_0\|}
\end{aligned}
\right\}_{t=1}^T
\end{equation}

\vspace{1em}
\paragraph{Orientation DMP:}
\begin{equation}
\text{DMP}_{\text{quat}}(q^{\text{new}}_0) = \left\{ y_t \;\middle|\;
\begin{aligned}
\tau \ddot{y}_t &= \alpha_z \left( \beta_z (\Delta q_T - y_t) - \tau \dot{y}_t \right) + f(s_t), \\
\tau \dot{y}_t &= \ddot{y}_t, \quad \tau \dot{s}_t = -\alpha_s s_t, \\
f(s) &= \frac{\sum_i \psi_i(s) w_i}{\sum_i \psi_i(s)} s\, \Delta q_T
\end{aligned}
\right\}_{t=1}^T
\end{equation}

\FloatBarrier
\subsection{Score Visualization}

\begin{figure}[h]
    \centering
    \includegraphics[width=0.75\linewidth]{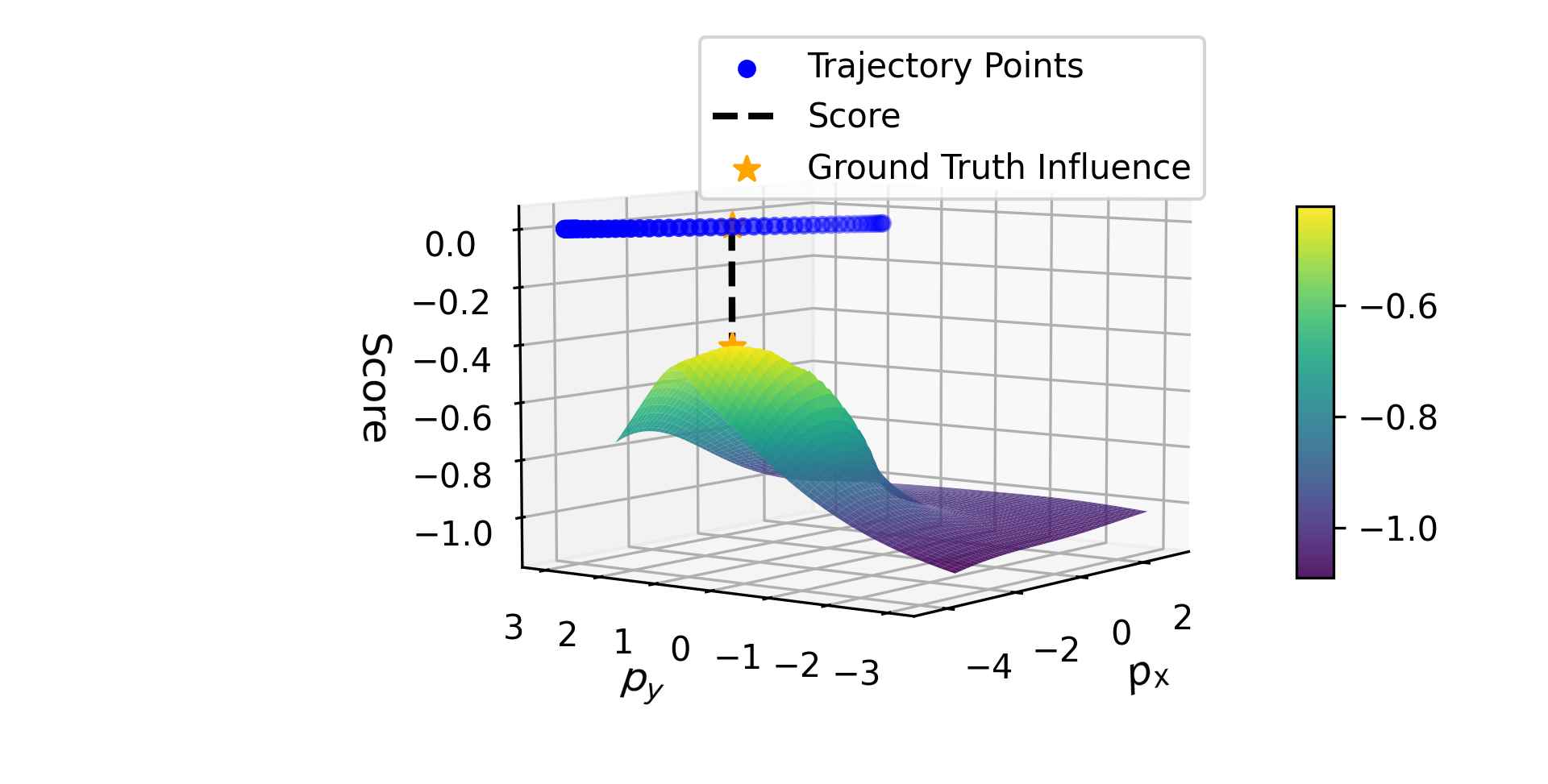}
    \caption{Score Landscape in 2D case for a trajectory length of $T = 50$. Notice the high score region around the ground-truth influence point.}
    \label{fig:high_score}
\end{figure}

\FloatBarrier
\subsection{Prompt to VLM}
\label{sec:prompt}
After we obtained the inferred influence point, it was projected on the initial RGB start state image. To extract semantic information about the task and the relevant environmental features, we adopt a two-phase querying strategy using a Vision-Language Model (VLM).

In the first phase, given the full demonstration trajectory overlaid on the start-state image, the VLM is queried to infer a concise task label that captures the primary objective of the motion. In the second phase, given the projected influence point on the same image and the previously inferred task label, the VLM is queried to identify the fine-grained environmental feature corresponding to the influence point and suggest an appropriate robot interaction location.

Few text-based examples are given to restrict the output to fine-grained details, without image-text pairs or in-context visual examples. The VLM is used solely as a tool for semantic extraction of task-relevant features, without any task-specific retraining or fine-tuning.

For the experiments conducted in this work, we employ GPT-4o~\citep{openai2024gpt4o}, one of the latest publicly available Vision-Language Models at the time of the experiments. Due to the rapid advancement of vision-language models, the proposed pipeline is designed to directly benefits from future models with improved visual reasoning capabilities. 

We provide the full text of the prompts used for querying the VLM in each phase below.

\paragraph{Phase 1: Task Label Inference} \begin{quote} \textbf{Instructions:} \ You are a motion understanding expert. \ Provided is an image showing the initial state of a robot task, overlaid with the robot's full demonstration trajectory. \

Based on the scene context and the projected trajectory, identify the task that the robot is attempting to perform. \

Respond with a concise task label that captures the primary objective of the robot's action. Focus solely on the relevant interaction. Ignore unrelated background objects. \

Examples of valid task labels include: ``place object on shelf,'' ``connect two components,''`` align tool with fixture,'' ``adjust object position,'' or ``slide object along surface.'' \

Your output should be a short phrase describing the task. \end{quote}

\paragraph{Phase 2: Fine-Grained Feature Identification} \begin{quote} \textbf{Instructions:} \ You are an environment reasoning expert. \ Provided is an image of the environment annotated with a projected spatial influence point, inferred from the robot's demonstration. \

The task is $\{\text{task label}\}$. Your goal is to reason about fine-grained, task-relevant environmental features based on the task and influence point. In particular:

\begin{itemize} \item Detect precise structural features that are critical for completing the task. \item Avoid vague descriptions like ``cabinet'' or ``box.'' Instead, refer to specific parts, boundaries, or interaction affordances. \end{itemize}

Based on the influence point and the task:

\begin{enumerate} \item Describe the fine-grained environmental feature that the projected point corresponds to that is related to the task. Be specific about the geometry, material, or function if relevant. \item Specify where the robot should grasp or interact with the object to successfully complete the task, using fine-grained features as references. Be specific about the geometry, material, or function if relevant.\end{enumerate}

Do not mention unrelated scene elements.

Provide two short phrases: one for the influence point description and one for the grasp location. \end{quote}

\paragraph{Example Outputs} \begin{itemize} \item Task label: \textit{Open cabinet door} \item Fine-grained description: \begin{itemize} \item (1) The projected point corresponds to the vertical seam between the cabinet door and the frame, near the edge where the door can be pulled open. \item (2) The robot should grasp the small metal door handle located along the seam to successfully open the cabinet door. \end{itemize} \end{itemize}

\FloatBarrier
\subsection{Real World Task Details and Failure Example Visualizations}

\begin{figure}[h]
    \centering
    \includegraphics[width=0.80\textwidth]{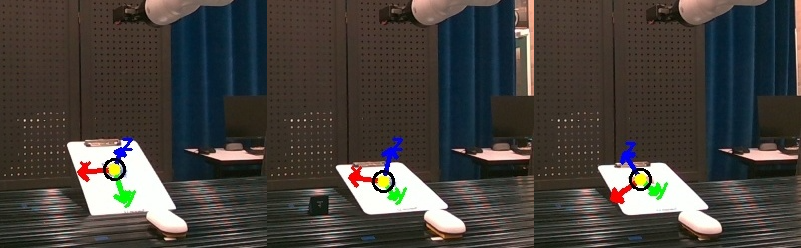}
    \caption{\textbf{Comparison of Extracted Local Frames in Surface Wiping Variants.} 
    (Left) Local frame extracted from the demonstration, correctly aligning the $z$-axis with the surface normal of a tilted board. 
    (Middle) Frame inferred for a successful execution on a flatter tilt, preserving the surface orientation and enabling stable wiping contact. 
    (Right) Frame inferred during a failed trial, where the $z$-axis does not follow the true surface normal, resulting in a wiping trajectory that drifts off the board or fails to maintain contact.}
    \label{fig:wiping_frame_comparison}
\end{figure}

\begin{figure}[t]
    \centering
    \begin{subfigure}[t]{0.48\textwidth}
        \centering
        \includegraphics[height=5cm]{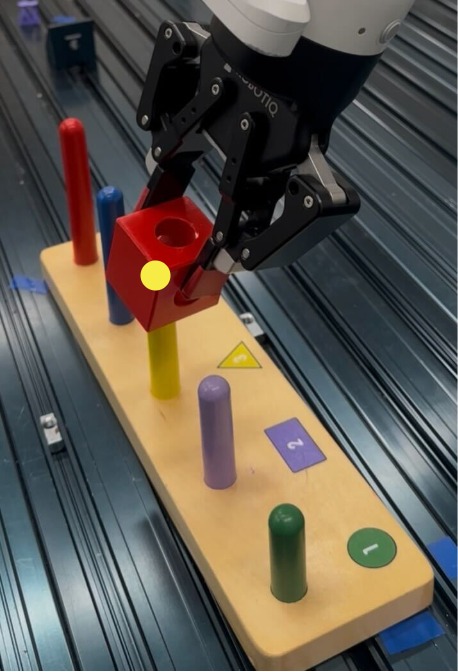}
        \caption{\textbf{Ours (DMP + Frame)} successfully hooks the red cube over the shorter yellow rod by adapting the trajectory downward before release.}
        \label{fig:ours_success}
    \end{subfigure}
    \hfill
    \begin{subfigure}[t]{0.48\textwidth}
        \centering
        \includegraphics[height=5cm]{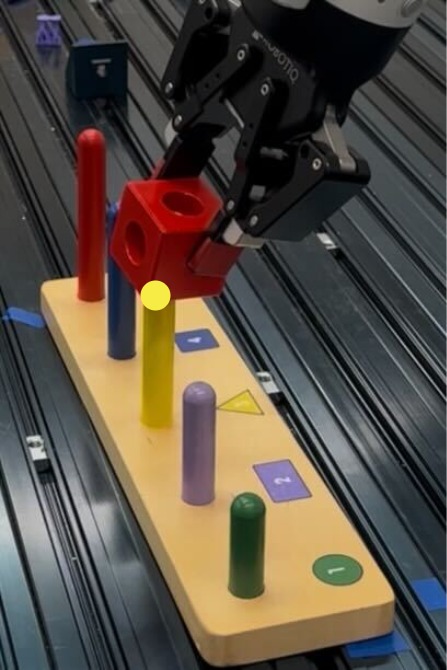}
        \caption{\textbf{Baseline DMP} fails to adapt to the rod height. Despite visual alignment, the cube misses the hook due to an insufficient downward motion.}
        \label{fig:baseline_failure}
    \end{subfigure}
    \caption{\textbf{Task Sensitivity to Rod Height Variation.} The yellow dot indicates the top of the yellow rod. A small height change leads to failure if the hook motion is not adapted. Our method infers a spatial reference frame that enables height-aware motion adjustment, while baseline DMP fails despite close visual alignment.}
    \label{fig:rod_height_comparison}
\end{figure}

\begin{figure}[t]
    
    \centering
    \includegraphics[width=0.80\textwidth]{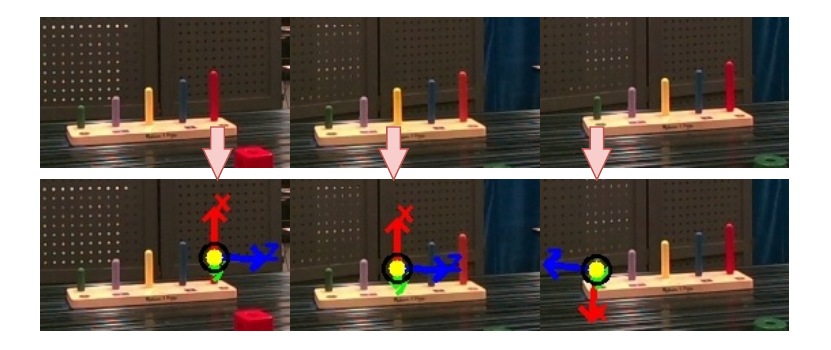}
    \caption{\textbf{Comparison of Extracted Local Frames in Drop Task Variants.} 
    (Left) Extracted local frame from the demonstration, aligned with the red rod. 
    (Middle) Inferred frame for a successful execution on the yellow rod, showing good alignment with the rod and consistent trajectory adaptation. 
    (Right) Inferred frame for a failed execution on the green rod, where the extracted frame is misaligned due to depth noise or segmentation error, leading to an incorrect drop trajectory.}
    \label{fig:drop_failure}
\end{figure}

\begin{figure}[t]
    \centering
    \includegraphics[width=0.95\textwidth]{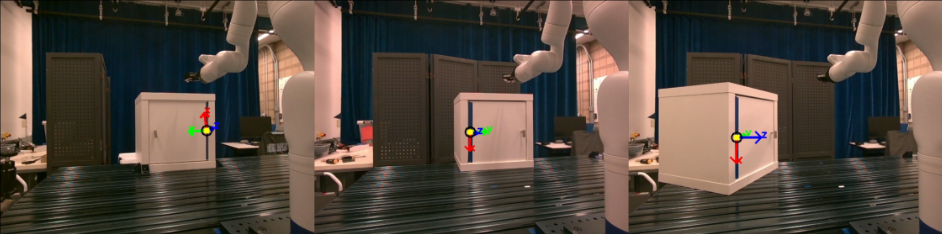}
    \caption{\textbf{Comparison of Extracted Local Frames in Cabinet Door Opening Variants.} 
    (Left) Extracted local frame from the demonstration, aligned with the door's hinge and surface in a frontal configuration. 
    (Middle) Frame inferred for a successful mirrored execution, correctly capturing the hinge orientation and aligning the $z$-axis with the surface normal. 
    (Right) Frame inferred in a failed mirrored execution, where the $z$-axis is not orthogonal to the surface, leading to an incorrect arc and failed opening trajectory.}
    \label{fig:door_frame_mirror}
\end{figure}

\end{document}